\documentclass[10pt]{article}

\usepackage[preprint]{main}

\usepackage{amsmath,amsfonts,bm}

\def\eqref#1{equation~\ref{#1}}

\def\1{\bm{1}}

\DeclareMathAlphabet{\mathsfit}{\encodingdefault}{\sfdefault}{m}{sl}
\SetMathAlphabet{\mathsfit}{bold}{\encodingdefault}{\sfdefault}{bx}{n}

\usepackage{amsmath, amssymb}

\usepackage{tabularx}
\usepackage{booktabs}
\usepackage{multirow}

\usepackage{graphicx}
\usepackage{xcolor}

\usepackage{subcaption}

\usepackage{algorithm}
\usepackage{algorithmic}

\usepackage{tikz}
\usetikzlibrary{
    positioning,
    shapes.geometric,
    arrows.meta,
    decorations.pathreplacing,
    calc
}

\usepackage{pgfplots}
\pgfplotsset{compat=1.17}

\usepackage{hyperref}
\usepackage{url}

\usepackage{pifont}

\title{Match or Replay: Self Imitating Proximal Policy Optimization}

\author{\name Gaurav Chaudhary  \email gauravch@iitk.ac.in \\
      \addr Department of Electrical Engineering\\
      Indian Institute of Technology Kanpur
      \and
      \name Laxmidhar Behera \email lbehera@iitk.ac.in\\
      \addr  Department of Electrical Engineering\\
      Indian Institute of Technology Kanpur
      \and
      \name Washim Uddin Mondal \email wmondal@iitk.ac.in \\
      \addr  Department of Electrical Engineering\\
      Indian Institute of Technology Kanpur
      }

\def\month{MM}  
\def\year{YYYY} 
\def\openreview{\url{https://openreview.net/forum?id=XXXX}} 

\begin{document}

\maketitle

\begin{abstract}
Reinforcement Learning (RL) agents often struggle with inefficient exploration, particularly in environments with sparse rewards. Traditional exploration strategies can lead to slow learning and suboptimal performance because agents fail to systematically build on previously successful experiences, thereby reducing sample efficiency.
To tackle this issue, we propose a self-imitating on-policy algorithm that enhances exploration and sample efficiency by leveraging past high-reward state-action pairs to guide policy updates. Our method incorporates self-imitation by using optimal transport distance in dense reward environments to prioritize state visitation distributions that match the most rewarding trajectory. In sparse-reward environments, we uniformly replay successful self-encountered trajectories to facilitate structured exploration.
Experimental results across diverse environments demonstrate substantial improvements in learning efficiency, including MuJoCo for dense rewards and the partially observable 3D Animal-AI Olympics and multi-goal PointMaze for sparse rewards. Our approach achieves faster convergence and significantly higher success rates compared to state-of-the-art self-imitating RL baselines. These findings underscore the potential of self-imitation as a robust strategy for enhancing exploration in RL, with applicability to more complex tasks.
\end{abstract}

\section{Introduction}
Deep Reinforcement Learning (DRL)~\citep{li2017deep} has achieved remarkable success in solving complex problems across a variety of domains, including robotic manipulation~\citep{han2023survey}, flight control~\citep{kaufmann2018deep}, epidemic control \citep{ling2024cooperating}, intelligent perception system~\citep{10097084}, and real-time strategy game-play~\citep{andersen2018deep}. However, despite these advancements, DRL algorithms still face significant challenges in efficient learning, leading to poor sample efficiency~\citep{baker2019emergent}. A major contributing factor is reliance on unguided exploration to discover near-optimal policies, leading to slow convergence, particularly in environments with sparse rewards.

Guided exploration using expert demonstration has been proposed as a potential solution. Approaches such as those in \citep{salimans2018learning, ecoffet2019go, xu2023improved, duan2017one, zhou2019watch, haldar2023watch} have used expert data to guide the agent's learning process. However, these methods often face challenges, such as difficulty obtaining expert demonstrations, the risk of bias, and the potential for convergence to suboptimal policies when demonstrations are insufficiently informative.

Our work addresses these challenges by exploiting the agent's past successful state-action transitions to guide exploration. To this end, we propose a novel Self-Imitation Learning (SIL)~\citep{oh2018self} approach that leverages the agent's past high-reward transitions to guide current policy updates—effectively ``bootstrapping" its learning process. This strategy enhances exploration and reduces the risk of divergence from effective behaviors~\citep{lin1992self}.
By leveraging the most rewarding past trajectories, our approach prevents the agent from deviating too far from previously learned successful behaviors, thereby improving exploration and learning efficiency. 

Self-imitation learning has been applied to diverse complex tasks such as robotics~\citep{Luo2023ReinforcementLI, luo2021self}, text-based games~\citep{Shi2023SelfimitationLF}, procedurally generated environments~\citep{Lin2023TakingCA}, interactive navigation~\citep{Kim2023VisualHS}, and large language models~\citep{Xiao2024HowTL}. Despite these advancements, a unified on-policy RL approach that effectively integrates self-imitation across both low-dimensional (structured) and high-dimensional (pixel-based) observations, while accommodating both dense and sparse reward settings, remains absent. Prior work, such as \citep{oh2018self}, has explored self-imitation in reinforcement learning; however, it primarily focuses on off-policy algorithms or incorporates off-policy elements when adapting self-imitation to on-policy methods like PPO \citep{schulman2017proximal}. In contrast, we introduce Self-Imitating Proximal Policy Optimization (SIPP)—a framework explicitly designed for on-policy RL that seamlessly integrates self-imitation into PPO’s update mechanism, without relying on replay buffers or off-policy corrections. By doing so, SIPP preserves PPO’s stability and theoretical guarantees while significantly enhancing exploration and sample efficiency.

For dense reward environments, we propose the MATCH strategy. Even with high-dimensional state and action spaces and dense rewards, RL agents struggle with sample efficiency, leading to slow convergence or suboptimal policies. As implemented in SIPP, self-imitation addresses this by guiding the agent toward high-value regions of the state space, thereby reducing divergence from effective policies and accelerating learning. Specifically, the MATCH strategy uses optimal transport~\citep{peyre2019computational}, particularly the Sinkhorn algorithm~\citep{cuturi2013sinkhorn}, to measure the similarity between the current policy's state distributions and the most rewarding episodic rollout from past experience. By prioritizing state-action transitions that closely match these rewarding distributions, the MATCH strategy ensures that exploration focuses on the self-encountered regions of the state space with high expected rewards.

For sparse and binary reward environments, we introduce the REPLAY strategy. Sparse and binary reward environments often present significant challenges for learning agents due to the limited availability of positive feedback. Our proposed REPLAY strategy, a variant of the MATCH strategy tailored to such sparse-reward scenarios, maintains an imitation buffer that stores previously encountered successful trajectories. However, while MATCH prioritizes individual state-action pairs, REPLAY introduces a trajectory-level replay mechanism tailored to sparse-reward settings. By replaying entire successful trajectories rather than isolated experiences, REPLAY improves learning in sparse-reward settings by repeatedly exposing the agent to high-return trajectories, which effectively reinforces key state-action dependencies over time.  Additionally, as demonstrated in the Animal-AI Olympics experiments, REPLAY can handle partial observability, making it more versatile than existing methods.

\noindent To summarize, our key contributions are as follows:

\begin{itemize}
\item Self-imitating on-policy algorithm: We propose Self-Imitating Proximal Policy Optimization (SIPP), a novel self-imitation learning algorithm that enhances exploration and sample efficiency in dense and sparse reward settings. 
\item Optimal transport-based prioritization: We introduce the MATCH strategy, which uses Optimal Transport~\citep{peyre2019computational} and the Sinkhorn algorithm~\citep{cuturi2013sinkhorn} to prioritize state-action transitions that closely match the state distribution of the most rewarding past episodic rollout, thereby improving learning efficiency in dense-reward environments.

\item Sparse reward scenario: We develop a REPLAY strategy that stores and reuses successful trajectories to reinforce long-term dependencies and improve learning from delayed rewards, resulting in enhanced sample efficiency.  
\item Diverse empirical validation: We validate SIPP through experiments across a wide range of environments, including complex MuJoCo~\citep{towers_gymnasium_2023} tasks, multi-goal PointMaze navigation~\citep{gymnasium_robotics2023github}, and partially observable 3D Animal-AI Olympics~\citep{crosby2019animal}, demonstrating significant improvements in learning efficiency and performance over state-of-the-art~\citep{oh2018self, gangwani2018learning}. 
\end{itemize}

\section{Related Work}
Many attempts have addressed the sample efficiency and exploration problem in reinforcement learning. However, this literature has divided the long work history mainly into guided and unguided exploration.

\textbf{Guided exploration} paradigms aim to leverage expert trajectories~\citep{chaudhary2025from, chaudhary2025moorl} to address RL agents' sample-efficiency and exploration problems. Recently, in this direction,~\citep{sontakke2024roboclip} presented an approach that uses a single demonstration and distilled knowledge contained in Video-and-Language Models (VLMs) to train a robotics policy. They use VLMs to generate rewards by comparing expert trajectories and policy rollouts. Another single demonstration guided approach was presented by \citep{libardi2021guided} for solving three-dimensional stochastic exploration. They exploit expert trajectories and value-estimate prioritized trajectories to learn optimal policies under uncertainty. Similarly, \citep{salimans2018learning} trained a robust policy using a single demonstration by replaying the demonstration for 
$n$ steps, after which agents learned in a self-supervised manner. To make the agent robust to randomness, they monotonically decrease the replay steps $n$. \citep{uchendu2023jump} presents an expert-guided learning. They employ two policies to solve tasks: the guide policy and the exploration policy. The guide policy introduces a curriculum of initial states for the exploration policy, significantly easing the exploration challenge and facilitating rapid learning. As the exploration policy becomes more proficient, the reliance on the guide policy diminishes, allowing the RL policy to develop independently and continue improving autonomously. This progressive reduction in the influence of the guide policy enables the agent to transition to a fully autonomous exploration phase, thereby enhancing its long-term performance and adaptability.

\citep{xu2023improved} uses expert demonstration to improve exploration in learning from demonstrations in sparse reward settings. They assign an exploration score to each demonstration, generate an episode, and train a policy to imitate exploration behaviors. \citep{nair2018overcoming} designs an auxiliary objective based on demonstrations to address hard exploration problems and gradually wean the demonstration guidance once the policy performs better than the demonstration. \citep{huang2023guided} used a two-component approach: a novel actor-critic-based policy-learning module that efficiently uses demonstration data to guide RL exploration, and a non-parametric module that employs nearest-neighbor matching and locally weighted regression for robust guidance propagation at states distant from the demonstrated ones.

\textbf{Unguided exploration} approaches use self-experience, count-based methods, or prioritized experience replay buffers to guide policy in hard exploration problems. In this literature, we focus only on approaches within the Self-Imitation Learning (SIL) paradigm, as coined by~\citep{oh2018self}. \citep{oh2018self} presented an approach for self-imitation learning for off-policy algorithms. They store experiences in a replay buffer and learn to imitate state-action pairs in the replay buffer only when the return in the past episode is greater than the agent’s value estimate. They also extended their approach to the on-policy algorithm. However, the proposed algorithm lacks a strong theoretical connection to on-policy algorithms.

\citep{gangwani2018learning} introduces the Stein Variational Policy Gradient (SVPG), a self-imitating algorithm designed for on-policy reinforcement learning. In this approach, policy optimization is framed as a divergence minimization problem, with the objective of minimizing the difference between the visitation distribution of the current policy and the distribution induced by experience replay trajectories with high returns. The method incorporates an auxiliary objective that regularizes this divergence, allowing for improved exploration and more effective policy updates. However, their experiments are limited to episodic, delayed, or noisy reward settings, which may restrict the generalizability of their results to more complex environments.

\citep{chen2020self} presents a SIL technique for off-policy algorithms. In their approach, they provide a constant reward at each step in addition to an episodic environment reward. Further, they maintain two replay buffers, one with 
$K$ highest episodic reward trajectories and the other with all agent-generated trajectories, and sample from these two replay buffers to train the policy. They limit their work to delayed episodic rewards. \citep{tang2020self} presents a self-imitation learning approach for off-policy learning by extending the traditional Q-learning with a generalized
n-step lower bound. They adopt SIL by leveraging trajectories where the behavior policy performs better than the current policy.
\citep{ferret2020self} proposes a self-imitating variant of DQN for dense reward environments. In this approach, they propose adopting self-imitation via a modified reward function. They augment the true reward with a weighted advantage term, the difference between a true discounted reward and an expected future return.

\citep{kang2020balancing} introduces the Explore-then-Exploit (EE) framework, which integrates Random Network Distillation (RND) \citep{burda2018exploration} and Generative Adversarial Self-Imitation Learning (GASIL) \citep{guo2018generative}. The framework addresses the exploration-exploitation trade-off by leveraging RND to facilitate exploration and prevent the policy from stagnating in local optima. At the same time, GASIL accelerates policy convergence by leveraging past successful trajectories. Rather than directly combining these methods, which could confuse the agent, the authors propose an interleaving approach in which the agent alternates between exploration and imitation based on specific criteria. 

Recently, \citep{li2023self} extended the SIL approach to Goal-Conditioned Reinforcement Learning. They achieve this by designing a target-action-value function that effectively combines the training mechanisms of the self-initiated policy and the actor policy. The SILP~\citep{luo2021self} method uses a planning mechanism for robotic manipulation that identifies effective policies from prior experience, enabling the agent to imitate high-quality actions even when explicit demonstrations are unavailable. By incorporating planning into the SIL framework, the agent can efficiently explore and exploit past successful behaviors. The approach improves the exploration-exploitation balance and enhances learning stability. 

The proposed approach aligns with unguided exploration with a focus on on-policy learning. The proposed approach uses past experiences to bootstrap policy learning, making a strong connection with the self-imitation learning paradigm.
\section{Preliminaries}

We consider a Markov Decision Process (MDP) symbolized as the tuple \(\mathcal{M}:\langle \mathcal{S}, \mathcal{A}, T, R,\gamma,\rho\rangle\) where $\mathcal{S}$ is the collection of environment states, $\mathcal{A}$ is the action space, \(T: \mathcal{S}\times \mathcal{A} \rightarrow 
 \Delta(\mathcal{S}) \) indicates the state transition function where $\Delta(\cdot)$ defines the probability simplex over its argument set, \(R : \mathcal{S}\times \mathcal{A}\times \mathcal{S} \rightarrow \mathbb{R}\) is the reward function, $\gamma$ $ \in $ $(0,1)$ denotes the discount factor, and $\rho\in \Delta(\mathcal{S})$ is the initial state distribution. At each time step $t$, the agent observes the state \(s_t\) and executes an action $a_t\in \mathcal{A}$. As a consequence, the state of the environment changes to $s_{t+1}$ following the transition law $T$, and the agent receives a reward $r_t=R(s_t, a_t, s_{t+1})$. A (stationary) policy is defined to be a map $\pi:\mathcal{S}\rightarrow \Delta(A)$. The reinforcement learning agent is trained to maximize the expected long-term discounted reward defined below over all $\pi\in \Pi$ where $\Pi$ is the collection of all policies~\citep{mondal2024improved}.
 \begin{align}
     J^{\pi}_{\rho} = \mathbf{E}_{\pi}\left[\sum_{t=0}^{\infty} \gamma^t r_t\big| s_0\sim \rho\right]
 \end{align}
 where $\mathbf{E}_{\pi}$ denotes expectation over all $\pi$-induced trajectories $\{(s_t, a_t)\}_{t=0}^{\infty}$ emanating from the initial distribution $\rho$. For large state spaces, the policies are represented by a parameter $\theta\in \mathbb{R}^{\mathrm{d}}$ where the dimension $\mathrm{d}$ is chosen such that $\mathrm{d}\ll |\mathcal{S}||\mathcal{A}|$. For Neural Network-based policies, $\theta$ is the weight parameter. In this framework, the agent's goal is to maximize $J_{\rho}^{\pi_{\theta}}\triangleq J(\theta)$ over $\theta\in \mathbb{R}^{\mathrm{d}}$. We have dropped the dependence of $\rho$ on $J(\theta)$ to simplify notations. We achieve this using a PPO-style gradient-based learning algorithm with few changes, driven by the self-imitation objective as explained in the forthcoming section.

\section{Method}
In this work, we propose an approach that guides an agent's exploration by combining its current behavior with past successful trajectories to enable on-policy RL. Unlike prior self-imitation methods that rely on off-policy data or modify the reward function, SIPP operates entirely within the on-policy framework. By modifying the rollout buffer sampling strategy (MATCH) or selectively replaying successful trajectories (REPLAY), SIPP reuses successful trajectories in a controlled fashion and integrates them into PPO’s on-policy training loop, without additional target networks or density-ratio corrections that are typical in off-policy RL. While this introduces some bias due to sample reuse, as discussed in works such as GePPO~\citep{queeney2021generalized}, empirical results indicate that the training remains stable and effective. This seamless integration with PPO ensures that our approach is both stable and efficient. The key highlights are as follows:

\begin{itemize}
\item Our approach does not alter the base RL policy (PPO) nor introduce new separate models requiring training, unlike prior works~\cite{gangwani2018learning, kang2020balancing}.
\item Our approach does not modify the true reward, preventing any bias in learning, unlike \citep{chen2020self}.
\item We address exploration by self-imitation for dense, sparse, and binary rewards encompassing both low and high-dimensional (pixel-based) observations, unlike \citep{oh2018self, gangwani2018learning}, which addressed only delayed and noisy rewards for low-dimensional observations in an on-policy setting.
\end{itemize}

\subsection{MATCH: A Self-Imitating Proximal Policy Optimization}

The MATCH strategy is designed to enhance exploration in dense reward environments by encouraging the agent to revisit transitions similar to those in the most rewarding past trajectory. Inspired by self-imitation learning, this approach assigns higher priority to state-action pairs that closely align with previously successful behaviors. To formally define the idea of similarity, we use \textit{Optimal Transport} (OT) as a principled method to compare empirical state visitation distributions.

Optimal Transport (OT)~\citep{cuturi2013sinkhorn, peyré2020computationaloptimaltransport, luo2023optimal} is a geometry-aware framework for comparing probability distributions. Suppose we are given two empirical distributions represented as:
\[
\mu = \frac{1}{T} \sum_{t=1}^{T} \delta_{x_t}, \quad \nu = \frac{1}{T'} \sum_{t'=1}^{T'} \delta_{y_{t'}},
\]
where \(x_t\) and \(y_{t'}\) denote sample points in a metric space (e.g., state embeddings), and \(\delta_{x_t}\) is the Dirac measure centered at \(x_t\), assigning unit mass at that point.
\begin{algorithm}[tb]
\caption{\textbf{MATCH}: Self-Imitating Proximal Policy}
\label{alg:ot}
    \begin{algorithmic}[1]
    \STATE \textbf{Input:} IET  $\xi$, initial state distribution $\rho$, batch size $B$, the inner loop length $H$, learning rate $\eta$    
    \STATE Initialize policy parameter $\theta_1 \gets \theta_0$\\
    \STATE Initialize imitation buffer $\mathcal{B_I} \leftarrow \{\}$\\
    \STATE Initialize data buffer $\mathcal{D} \leftarrow \{\}$\\
    \textcolor{blue}{**Outer Loop**}
        \FOR {$k\in \{1, 2, \cdots\}$}
        \STATE Collect $\pi_{\theta_k}$-induced trajectory $\{(s_t^k, a_t^k)\}_{t=1}^T$ in $\mathcal{D}$ 
        \STATE Update $\mathcal{B_I}$ by storing the highest-rewarding trajectory seen so far.
        \STATE Obtain advantage estimates $\hat{A}_1,\cdots,\hat{A}_T$ corresponding to $\pi_{\theta_k}$and the state-action pairs of the trajectory in $\mathcal{D}$\\
        \STATE $\theta_{k, 0}\gets \theta_k$, $\theta_{k, -1}\gets \theta_{k-1}$\\
        \textcolor{blue}{**Inner Loop**}
        \FOR{$h\in \{0, \cdots, H-1\}$}
        \STATE Sample a $B$-sized batch of state-actions from $\mathcal{D}$ either uniformly or weighted by \eqref{otd}, controlled by $\xi$
        \STATE Update $\theta_{k, h}$ following \eqref{eq_washim_theta_kh_update}
        \ENDFOR
        \STATE $\theta_{k+1}\gets \theta_{k, H}$
        \STATE Empty data buffer $\mathcal{D} \leftarrow \{\}$
        \ENDFOR 
    \end{algorithmic}
\end{algorithm}
The squared Wasserstein distance between \(\mu\) and \(\nu\) is given by:
\begin{equation}
W^2(\mu, \nu) = \min_{\zeta \in \Gamma} \sum_{t=1}^{T} \sum_{t'=1}^{T'} c(x_t, y_{t'}) \zeta_{t t'},
\end{equation}
where \(\Gamma = \left\{ \zeta \in \mathbb{R}_{+}^{T \times T'}: \zeta \mathbf{1}_{T'} = \frac{1}{T} \mathbf{1}_{T}, \zeta^\top \mathbf{1}_{T} = \frac{1}{T'} \mathbf{1}_{T'} \right\}\) is the set of doubly stochastic coupling matrices, and \(c(x_t, y_{t'})\) is the cost of transporting unit mass from \(x_t\) to \(y_{t'}\). We invoke the cosine distance as the cost. To solve the above optimization efficiently, we apply the Sinkhorn algorithm~\citep{cuturi2013sinkhorn}, which incurs a computational complexity of \(\mathcal{O}(T T')\).

The MATCH algorithm has a nested loop structure. At the $k$th instance of the outer loop, it produces a $T$-length trajectory $\{(s_t^k, a_t^k)\}_{t=1}^T$ produced by the current policy $\pi_{\theta_k}$ (based on the task, $T$ is either deterministic or random) and stores it in a data buffer $\mathcal{D}$. If the trajectory stored in $\mathcal{D}$ is the highest-rewarding one seen so far, then it is also stored in the imitation buffer $\mathcal{B}_{\mathcal{I}}$, replacing any earlier-stored trajectories in it. Let $\{s_1^k, \cdots, s_T^k\}$ and $\{s_1^e, \cdots, s_{T'}^e\}$ be the states of the trajectories stored in $\mathcal{D}$ and $\mathcal{B}_{\mathcal{I}}$ respectively. Their empirical distributions are given as:
\begin{align}
\hat{q}_k = \frac{1}{T} \sum_{t=1}^{T} \delta_{s^k_t}, \quad \hat{q}_e = \frac{1}{T'} \sum_{t'=1}^{T'} \delta_{s^e_{t'}},
\end{align}
Their corresponding OT distance is given as:
\begin{equation}
W^2(\hat{q}_\pi, \hat{q}_e) = \min_{\zeta \in \Gamma} \sum_{t=1}^{T} \sum_{t'=1}^{T'} c(s^k_t, s^e_{t'}) \zeta_{t t'}.
\end{equation}

Let $\zeta^*$ be the solution to the above optimization. We define an OT-based similarity score for each state \(s^k_t\) in the trajectory in $\mathcal{D}$ as follows.
\begin{equation}
d_{\mathrm{OT}}(s^k_t) = -\sum_{t'=1}^{T'} c(s^k_t, s^e_{t'}) \zeta^*_{t t'}
\label{otd}
\end{equation}

The inner loop starts with initialization: $\theta_{k, 0}\leftarrow \theta_k$. At the $h$th instant of the inner loop, the agent chooses a batch of state-actions $\{(s_j^k, a_j^k)\}_{j\in \mathcal{J}}$ from $\mathcal{D}$ either via a uniform probability (exploration) or a priority-based strategy determined by the similarity score in \eqref{otd} (imitation). The choice between exploration and imitation is decided by a hyperparameter $\xi$, called the Imitation-Exploration Trade-off (IET) coefficient. For the chosen batch of state-action pairs, we can now define the PPO-based surrogate loss function as follows \citep{schulman2017proximal}.
\begin{equation}
L^{\text{PPO}}(\theta_{k,h}) = \mathbb{E}_j \left[ \min\left( r_j^{kh} \, A_j, \; \text{clip}\left( r_j^{kh}, 1 - \epsilon, 1 + \epsilon \right) A_j \right) \right],
\label{ppo_loss}
\end{equation}
where $\mathbb{E}_j$ denotes the empirical average over $j\in \mathcal{J}$, $\epsilon$ defines a clipping hyperparameter, $A_j$ is the estimate of the advantage~\citep{schulman2015high} function corresponding to the policy $\pi_{\theta_{kh}}$ and the pair $(s_j, a_j)$, and the ratio $r_j^{kh}$ is given as follows. 
\begin{align*}
    r_j^{kh} = \dfrac{\pi_{\theta_{k,h}}(a_j|s_j)}{\pi_{\theta_{k, h-1}}(a_j|s_j)}
    \label{eq_washim_theta_kh_update}
\end{align*}
where we use the convention that, for $h=0$, $\theta_{k,h-1}\gets \theta_{k-1}$. The policy parameter is now updated using gradient descent.
\begin{align}
    \theta_{k, h+1} \gets \theta_{k, h} - \eta \nabla_{\theta} L^{\text{PPO}}(\theta_{k, h})
\end{align}
where $\eta$ is the learning rate. Finally, we assign $\theta_{k+1}\gets \theta_{k, H}$ where $H$ is the inner loop length and start the $(k+1)$th instant of the outer loop. Algorithm \ref{alg:ot} summarizes the entire process.

Note that our proposed approach does not rely on any explicit expert policy. Instead, it utilizes the distribution of the most-rewarding trajectory generated by the current behavior policy (stored in the imitation buffer) as the surrogate expert policy. This removes the dependency on external expert trajectories and leverages the agent's high-performing experiences. Our algorithm ensures that the imitation buffer evolves continuously as the agent discovers better-performing trajectories, thus adapting the surrogate expert distribution over time.

\begin{algorithm}[tb]     \caption{\textbf{REPLAY}: Self-Imitating Proximal Policy}
    \label{alg:AGE}
    \begin{algorithmic}[1]
        \STATE \textbf{Input:} IET $\xi$, initial state distribution $\rho$, batch size $B$, the learning rate $\eta$, imitation buffer length $L$
        \STATE Initialize policy parameter $\theta_1\gets \theta_0$\\
        \STATE Initialize imitation buffer $\mathcal{B_I} \leftarrow \{\}$\\
        \STATE Initialize data buffer $\mathcal{D} \leftarrow \{\}$
    
        \FOR{$k\in \{1, 2, \cdots\}$}
        \STATE Sample $\tau\sim \mathrm{Bernoulli}(\xi)$
        \IF{$\tau=0$}
        \STATE Sample a trajectory from $\mathcal{B_I}$ and store it in $\mathcal{D}$ 
        \ELSIF{$\tau=1$}
        \STATE Store a $\pi_{\theta_k}$-induced trajectory in $\mathcal{D}$ 
        \ENDIF\\
        \STATE Obtain advantage estimates  $\hat{A}_1,\cdots,\hat{A}_T$ corresponding to $\pi_{\theta_k}$
        and the state-action pairs of the trajectory in $\mathcal{D}$
        \STATE Sample a $B$-sized batch of state-actions $\{(s_j^k, a_j^k)\}_{j\in \mathcal{J}}$ from data buffer $\mathcal{D}$ with uniform probability\\
        \STATE Update $\theta_k$ following \eqref{eq_washim_grad_descent_replay}\\
        \STATE Update $\mathcal{B_I}$ by storing the highest $L$ rewarding trajectories seen so far\\
        \STATE Empty data buffer $\mathcal{D} \leftarrow \{\}$
        
        \ENDFOR
    \end{algorithmic}
\end{algorithm}

\subsection{REPLAY: Self-Imitating Proximal Policy} 
This section addresses the exploration challenge in sparse-reward settings using self-imitation. Sparse-reward environments often pose significant challenges for learning agents due to the limited availability of positive feedback. To mitigate this challenge, we propose the REPLAY strategy, adapted from~\citep{libardi2021guided}, a variant of the MATCH strategy tailored specifically for sparse-reward scenarios. Unlike MATCH, which uses past trajectories to generate preferences for current trajectories, the REPLAY strategy focuses on directly replaying successful past behaviors. 

The structure of REPLAY is very similar to that of MATCH, except for some modifications mentioned below. REPLAY also maintains an imitation buffer, $\mathcal{B}_{\mathcal{I}}$, and a data buffer, $\mathcal{D}$. However, unlike MATCH, in this case, $\mathcal{B}_{\mathcal{I}}$ dynamically stores multiple trajectories. The capacity of $\mathcal{B}_{\mathcal{I}}$ is a given hyperparameter. REPLAY runs in multiple epochs. Let the policy parameter at the $k$th epoch be $\theta_k$. The agent either selects a trajectory from $\mathcal{B}_{\mathcal{I}}$ at random (imitation) or generates a trajectory induced by the current policy $\pi_{\theta_k}$ (exploration), and stores it into the data buffer $\mathcal{D}$. The probability of choosing either of these events is determined by the IET parameter $\xi$. This sampling mechanism ensures a balance between imitation and exploration. A higher value of IET emphasizes exploitation by prioritizing trajectory sampling from $\mathcal{B_I}$, while a lower value encourages exploration by focusing on the trajectories generated from the agent's most recent interactions.

Next, a batch of state-action pairs, $\{(s^k_j, a^k_j)\}_{j\in \mathcal{J}}$ of length $B$, is uniformly selected from the trajectory in $\mathcal{D}$. We can now define the surrogate PPO objective $L^{\text{PPO}}(\theta_k)$ in a way similar to \eqref{ppo_loss}. The policy parameter $\theta_k$ is updated via gradient descent.
\begin{align}
    \theta_{k+1} \gets \theta_k -\eta \nabla_{\theta} L^{\text{PPO}}(\theta_k)
    \label{eq_washim_grad_descent_replay}
\end{align}
where $\eta$ is the learning rate. Finally, the data buffer is emptied, the imitation buffer $\mathcal{B}_{\mathcal{I}}$ is updated by storing the top $L$ number of rewarding trajectories seen so far, where $L$ is the capacity of $\mathcal{B}_{\mathcal{I}}$, and the agent is prepared for the $(k+1)$th update epoch. Observe that the update procedure of $\mathcal{B}_{\mathcal{I}}$ resembles the first-in first-out (FIFO) queuing mechanism. Algorithm \ref{alg:AGE} summarizes the entire process.

In our algorithm, the process of ``replay'' refers to including the trajectories stored in the imitation buffer $\mathcal{B_I}$ into the data buffer $\mathcal{D}$ and treating them as potential behaviors generated by the current policy. This ensures that the agent encounters successful trajectories at each epoch to guide policy learning, even in sparse-reward environments where most interactions yield little to no reward. This selective reuse of high-rewarding trajectories improves the sample efficiency and reduces the risk of the agent getting stuck in 
unproductive exploration loops. Our REPLAY strategy thus offers a robust framework for learning in sparse-reward environments.
\begin{figure*}[t]
\centering
\subfloat{\label{fig:im11}}
\subfloat{\label{fig:im1}}
\subfloat{\label{fig:im2}}
\subfloat{\label{fig:im3}}
\subfloat{\label{fig:im4}}
\subfloat{\label{fig:im6}}
\subfloat{\label{fig:im7}}
\subfloat{\label{fig:im9}}
\subfloat{\label{fig:im10}}
\begin{tikzpicture}
\node (im11) [xshift =-83ex,yshift = 8ex]{\includegraphics[scale=0.22]{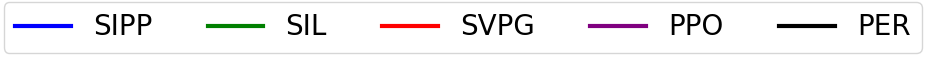}};
\node (im1) [xshift=-125ex, yshift = -5ex]{\includegraphics[scale=0.25]{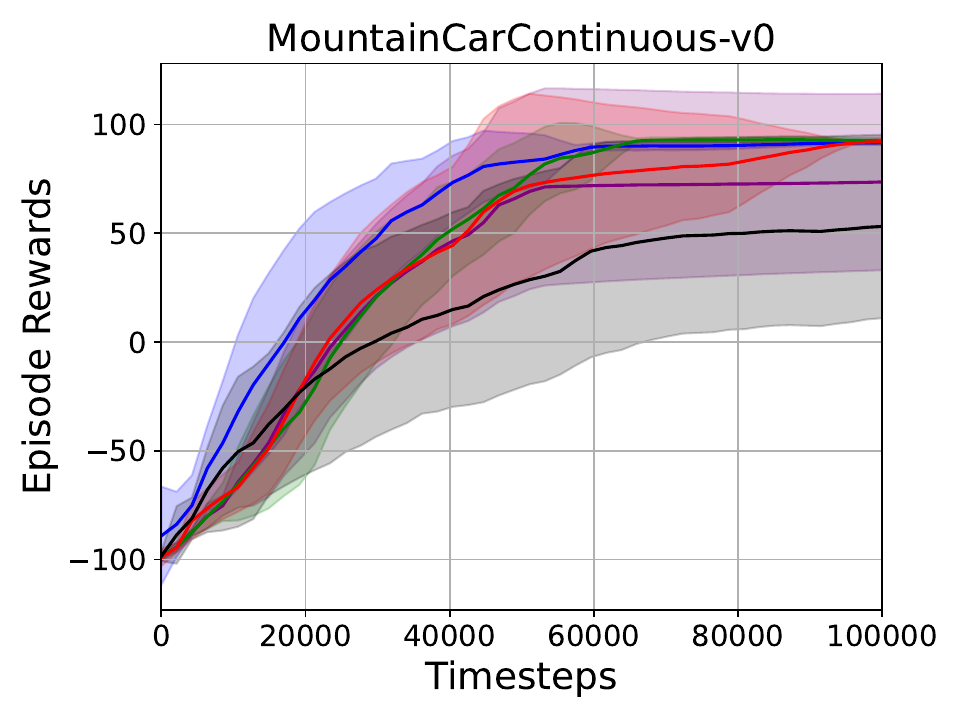}};
\node (im2) [xshift=-98.5ex,yshift = -5ex]{\includegraphics[scale=0.25]{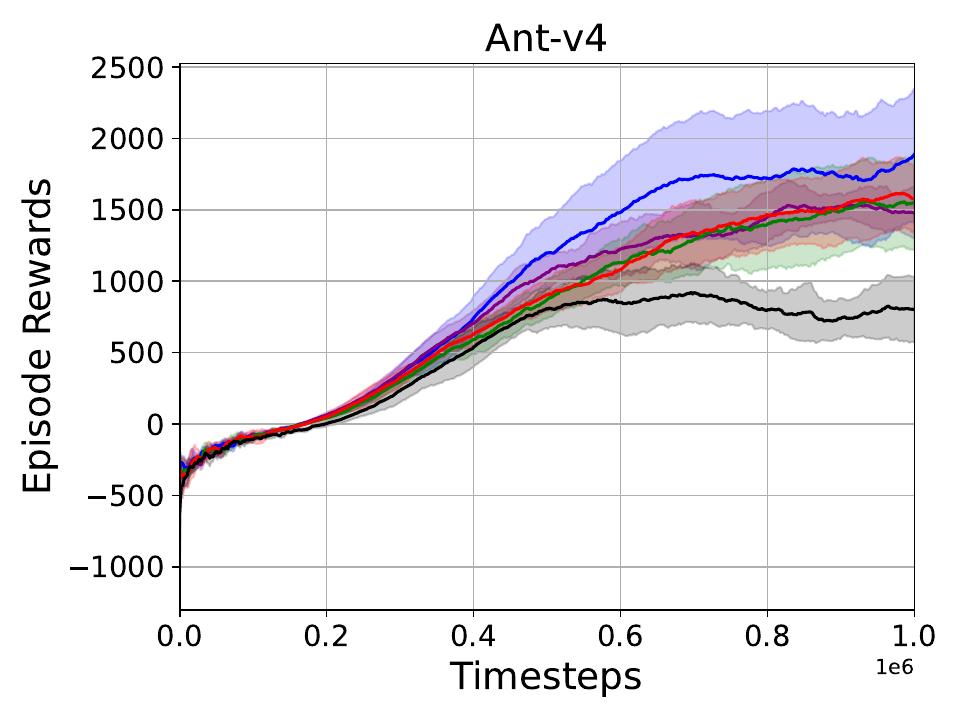}};
\node (im3) [xshift=-70.5ex,yshift = -5ex]{\includegraphics[scale=0.25]{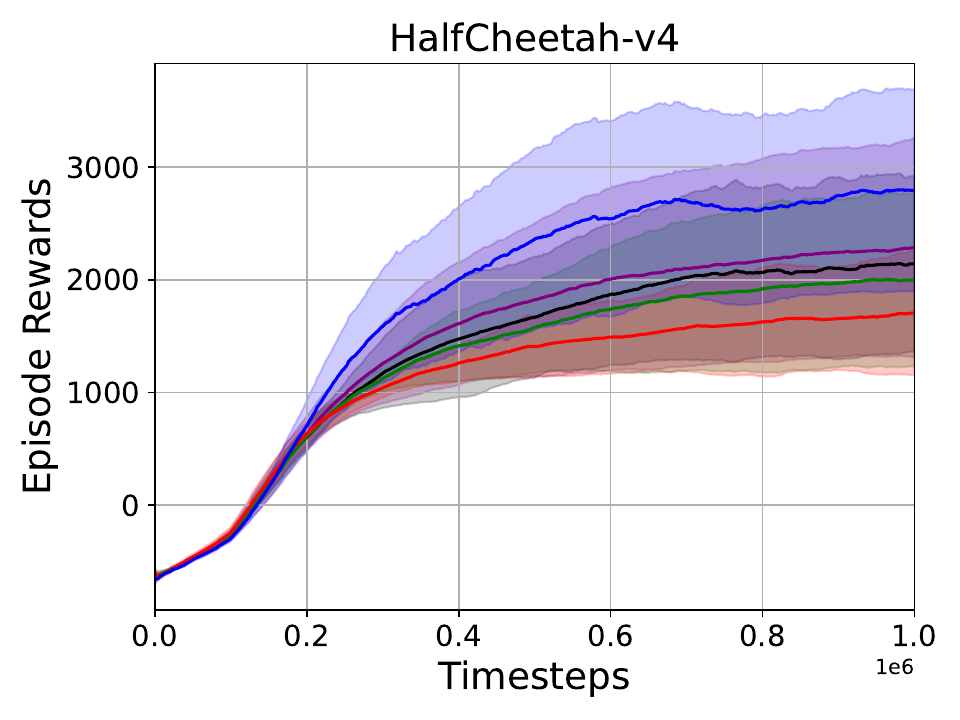}};
\node (im4) [xshift=-42.5ex, yshift = -5ex]{\includegraphics[scale=0.25]{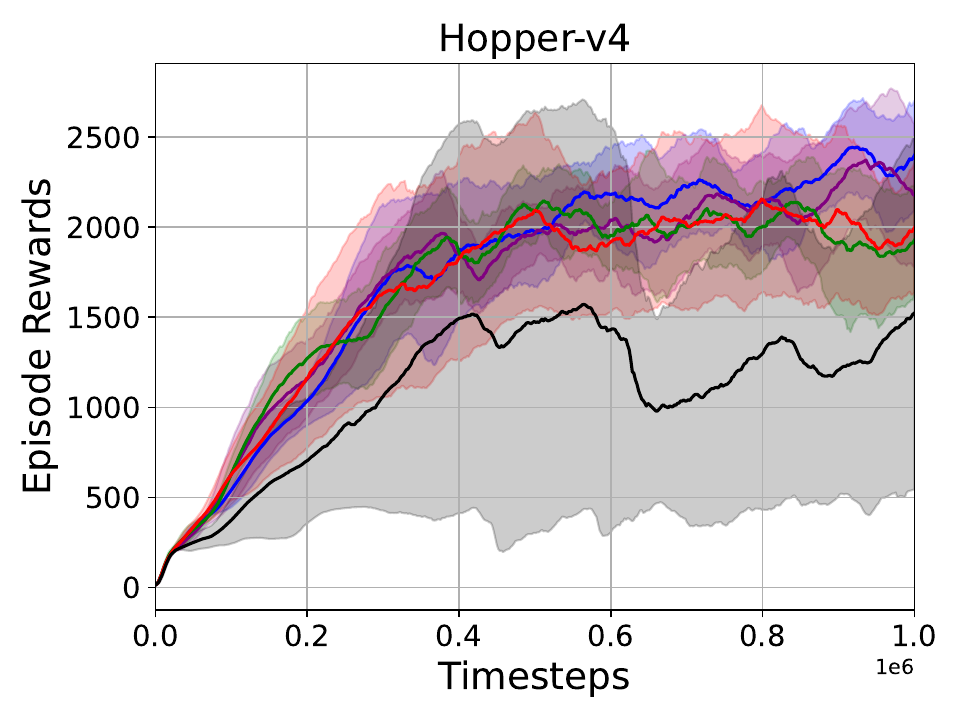}};
\node (im6) [xshift=-125ex, yshift=-28ex]{\includegraphics[scale=0.25]{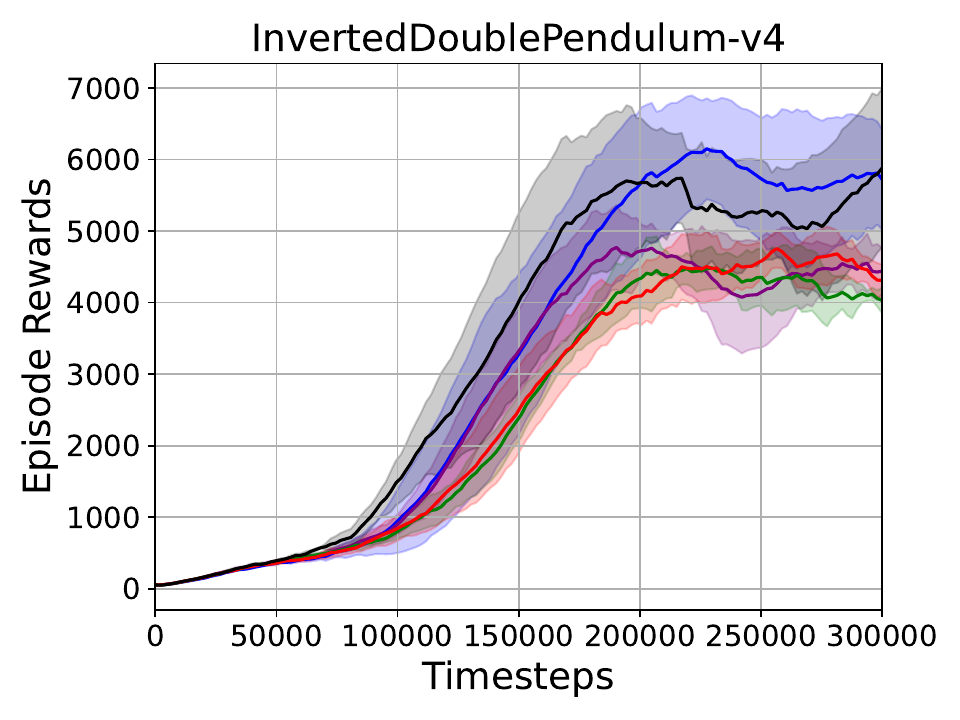}};
\node (im7) [xshift=-98.5ex, yshift=-28ex]{\includegraphics[scale=0.25]{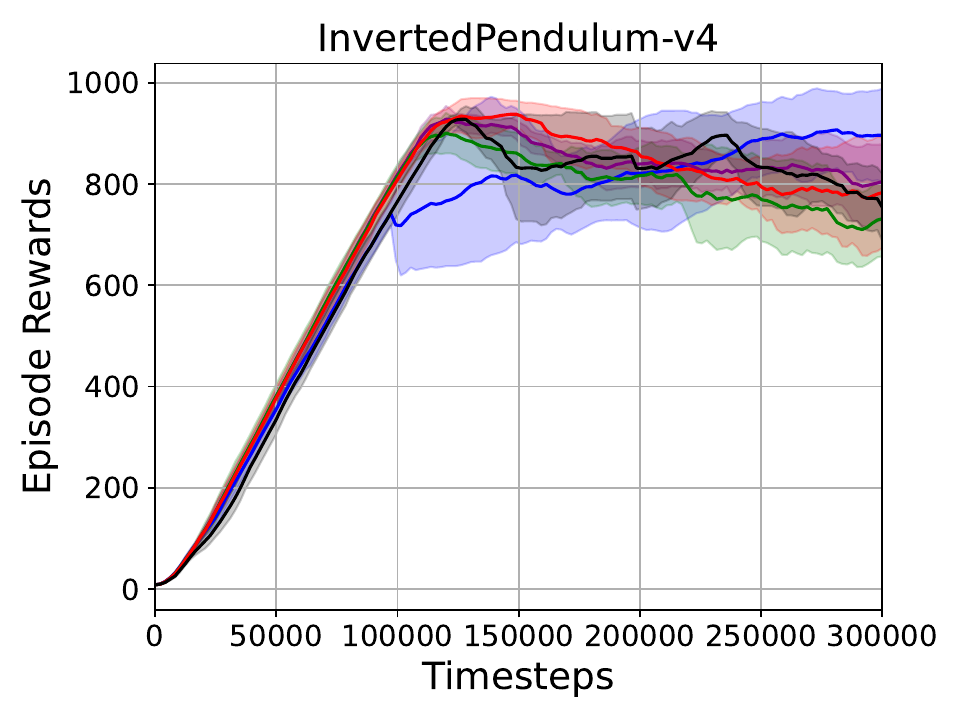}};
\node (im9) [xshift=-70.5ex, yshift=-28ex]{\includegraphics[scale=0.25]{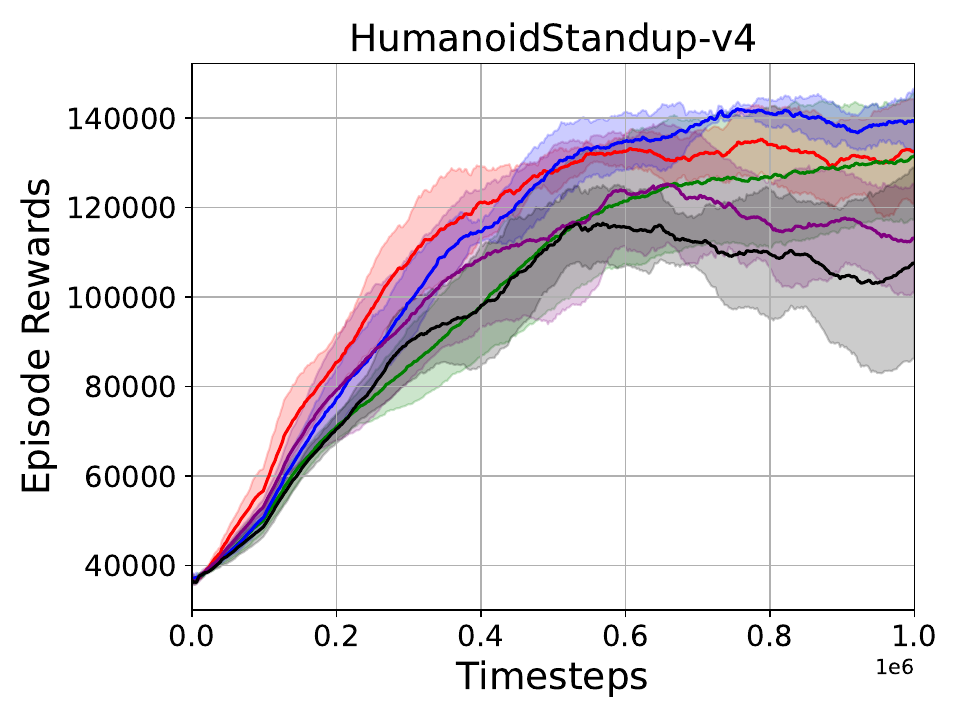}};
\node (im10) [xshift=-42.5ex, yshift=-28ex]{\includegraphics[scale=0.25]{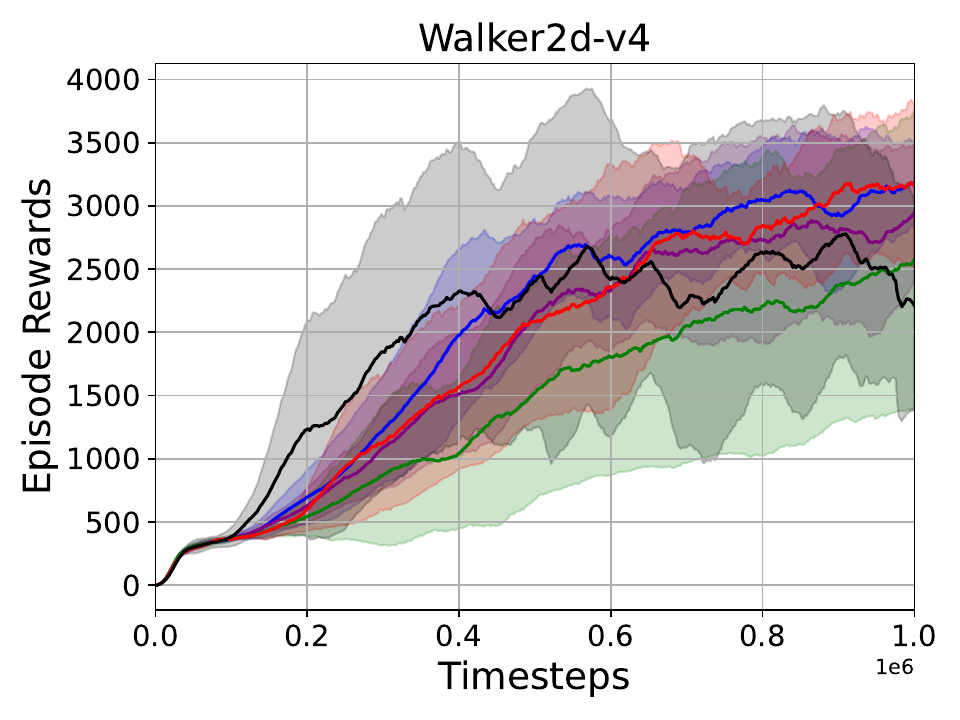}};
\end{tikzpicture}
\caption{Results show the performance of 8 MuJoCo~\citep{towers_gymnasium_2023} continuous control tasks (refer to Figure~\ref{fig:SIPP-match} for results on all tasks). The plots show the learning curves and the episodic rewards along the y-axis, evaluated under the current policy. The reported results are the mean across seven seeds, with shaded regions indicating the standard deviation. The proposed algorithms outperform all baselines across all tasks, achieving competitive or better performance.}
\label{fig:SIPP-match-1}
\end{figure*}

\section{Experiments}
In this section, we aim to answer the following questions:
\begin{itemize}
    \item Does bootstrapping policy learning, with its few past experiences, enhance sample efficiency and hard exploration across diverse tasks?
    \item Can a single past successful behavior be sufficient to guide policy learning in complex sequential continuous control tasks?
    \item Is replaying past successful trajectories sufficient for policy learning in multi-goal and partially observable, sparse reward settings? 
\end{itemize}
\subsection{Implementation Details}
For dense reward environments in MuJoCo~\citep{towers_gymnasium_2023}, we implement the \textit{Match} strategy. The network architecture utilizes a multi-layer perceptron (MLP) with two hidden layers containing 64 units and $tanh$ as the activation function. The PPO policy is updated over 10 epochs per training iteration. Training batches are sampled uniformly or prioritized based on the optimal transport distance between the current trajectories and past best episodic rollouts, controlled by the Imitation-Exploration Trade-off coefficient $\epsilon$ at each epoch. The imitation buffer is initiated with size 1. Further details about the hyperparameters and implementation can be found in the supplementary material Table~\ref{table1}.

For sparse-reward settings, such as multi-goal PointMaze~\citep{gymnasium_robotics2023github} navigation tasks, we adopt the \textit{Replay} strategy. This setup uses the same MLP-based architecture as MuJoCo environments. Similarly, for the Animal-AI Olympics~\citep{crosby2019animal}, a partially observable 3D environment with binary rewards, we apply the \textit{Replay} strategy but with a different architecture: a three-layer convolutional neural network (CNN). The input comprises the last four stacked frames (84 × 84 RGB pixels). In both PointMaze and Animal-AI tasks, the rollout buffer is populated with trajectories sampled either from the current policy with probability
$1-\epsilon$ or from the imitation buffer $\mathcal{B_I}$ with probability $\epsilon$. The imitation buffer is initialized with size 10. Comprehensive details of the hyperparameters for all environments are provided in the supplementary material Table~\ref{table2}.
\subsection{Choice of Baselines:}
Self-imitation learning (SIL) has been primarily explored to enhance exploration in off-policy reinforcement learning (RL) algorithms, with limited focus on on-policy RL algorithms. Further, recent works~\citep{Luo2023ReinforcementLI, Xiao2024HowTL, Shi2023SelfimitationLF} predominantly focus on problem-specific adoption of SIL rather than advancing SIL from a broader algorithmic perspective. Notably, there remains limited analysis of SIL's potential for on-policy RL algorithms across diverse problem settings, leaving a significant gap in understanding its general applicability. This limits the choice of baseline for our problem.

\textbf{PPO:} PPO is vanilla proximal policy algorithm~\citep{schulman2017proximal}, which does not impose any self-imitation learning paradigm.

\textbf{SIL-PPO:} SIL~\citep{oh2018self} is an off-policy RL algorithm that imitates past state-action pairs that have higher returns than agent value estimates. The proposed approach was also extended to the on-policy PPO~\citep{schulman2017proximal} algorithm with a focus on dense or delayed rewards. Further, as highlighted by~\citep{oh2018self}, SIL lacks a theoretical connection with on-policy algorithms. 

\textbf{SVPG-PPO:} SVPG~\citep{gangwani2018learning} is a self-imitating on-policy algorithm that uses Stein variational gradient descent to minimize the divergence between the current policy’s visitation distribution and that of past high-return trajectories. Unlike SIPP, SVPG introduces an auxiliary objective that regularizes this divergence, potentially complicating the learning process.

\textbf{PER-PPO}: Prioritize Experience Replay (PER)~\citep{schaul2015prioritized} technique uses TD-error based transition prioritization. We extended this method to PPO, prioritizing samples in the rollout buffer based on TD error. We use a strategy similar to our method to balance exploration and exploitation.

\subsection{Performance of \textit{Match} on Continuous Control Tasks}
In this section, we investigate the effect of self-imitation on continuous control tasks with dense rewards. We evaluate the performance of our \textit{SIPP-Match} strategy across 10 MuJoCo~\citep{towers_gymnasium_2023} tasks, using selected baselines.

Compared with all the baselines, the performance of the \textit{Match} strategy on continuous control tasks is shown in Figure~\ref{fig:SIPP-match-1}. The proposed \textit{Match} algorithm outperforms PPO~\citep{schulman2017proximal} and SIL-PPO across all tasks, with SVPG+PPO lagging in most tasks, except for competitive performance on Walker2d-v4 and Humanoid-v4. PER~\citep{schaul2015prioritized} uses a TD-error-based strategy that prioritizes transitions and lags across all tasks using value function estimates.

In MuJoCo benchmark environments, the agent benefits from continuous feedback via a smooth and dense reward structure, facilitating faster exploration and learning. Despite this, our experiments demonstrate that the optimal transport distance, prioritizing self-bootstrapping, can further enhance exploration for the proximal policy. 
By prioritizing the most informative experiences, our method ensures that the agent focuses on high-value learning opportunities, accelerating convergence and improving policy robustness.

Additionally, the proposed approach stores only the states visited by the most rewarding past episodic rollout, making it simpler than prior methods. Unlike our approach,~\citep{oh2018self} compares returns of past experiences with agent value estimates to select experiences for self-imitation, which can be noisy and introduce bias in policy learning ~\citep{libardi2021guided,raileanu2021decoupling}. Furthermore, \citep{gangwani2018learning} uses Stein variational gradient descent as a regularizer to minimize divergence between state-action visitation distributions of the current policy and past rewarding experiences. However, their approach introduces bias in policy learning, which they address by simultaneously learning multiple diverse policies.

In summary, the proposed~\textit{Match} algorithm integrates seamlessly with PPO without introducing additional learning parameters and requires only one trajectory to guide self-imitation learning. It introduces a single hyperparameter, IET coefficient ($\epsilon$), to control whether training batches are uniformly sampled or prioritized using optimal transport distance based on past successful episodic rollouts.
The single hyperparameter provides a simple mechanism to balance exploration and exploitation. This approach offers a practical and efficient solution to enhance reinforcement learning performance in complex environments.

\subsection{Performance of \textit{Replay} in Sparse Reward Tasks}
In this section, we empirically evaluate self-imitation performance in sparse and binary reward settings. We believe that self-imitation can play a crucial role in such reward settings, as the ability of an agent to reach some success can be extremely difficult with sparse rewards. Previous work was limited to MuJoCo environments with dense or delayed rewards. Motivated by this, we evaluate the performance of \textit{SIPP-Replay} on a diverse set of tasks, including multi-goal gymnasium-robotics PointMaze navigation sparse reward environments~\citep{gymnasium_robotics2023github}  and partially observable 3-dimensional Animal-AI Olympics ~\citep{crosby2019animal} binary reward environments. 
\begin{figure*}[t]
\centering
\subfloat{\label{fig:ima1}}
\subfloat{\label{fig:ima2}}
\subfloat{\label{fig:ima3}}
\subfloat{\label{fig:ima4}}
\subfloat{\label{fig:ima5}}
\begin{tikzpicture}
\node (im1) [xshift=-120ex, yshift=0ex]{\includegraphics[scale=0.075]{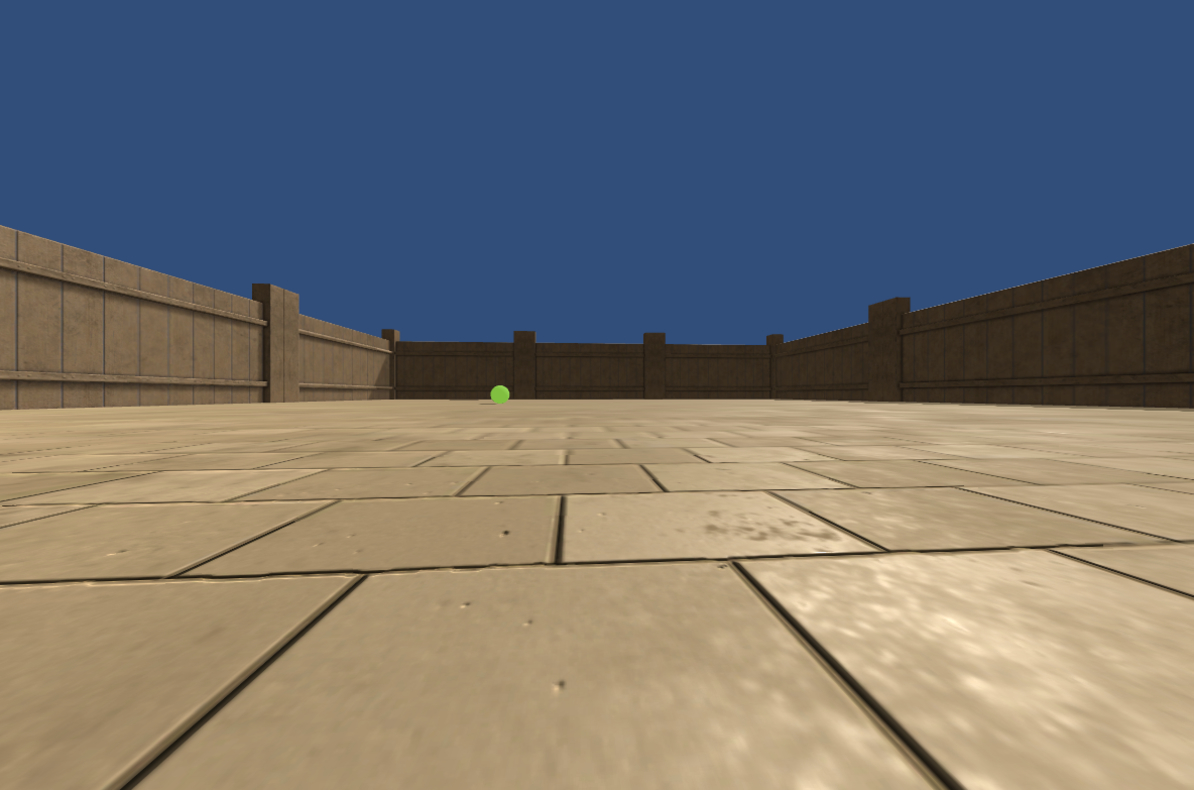}};
\node (im2) [xshift=-98ex,yshift=0ex]{\includegraphics[scale=0.075]{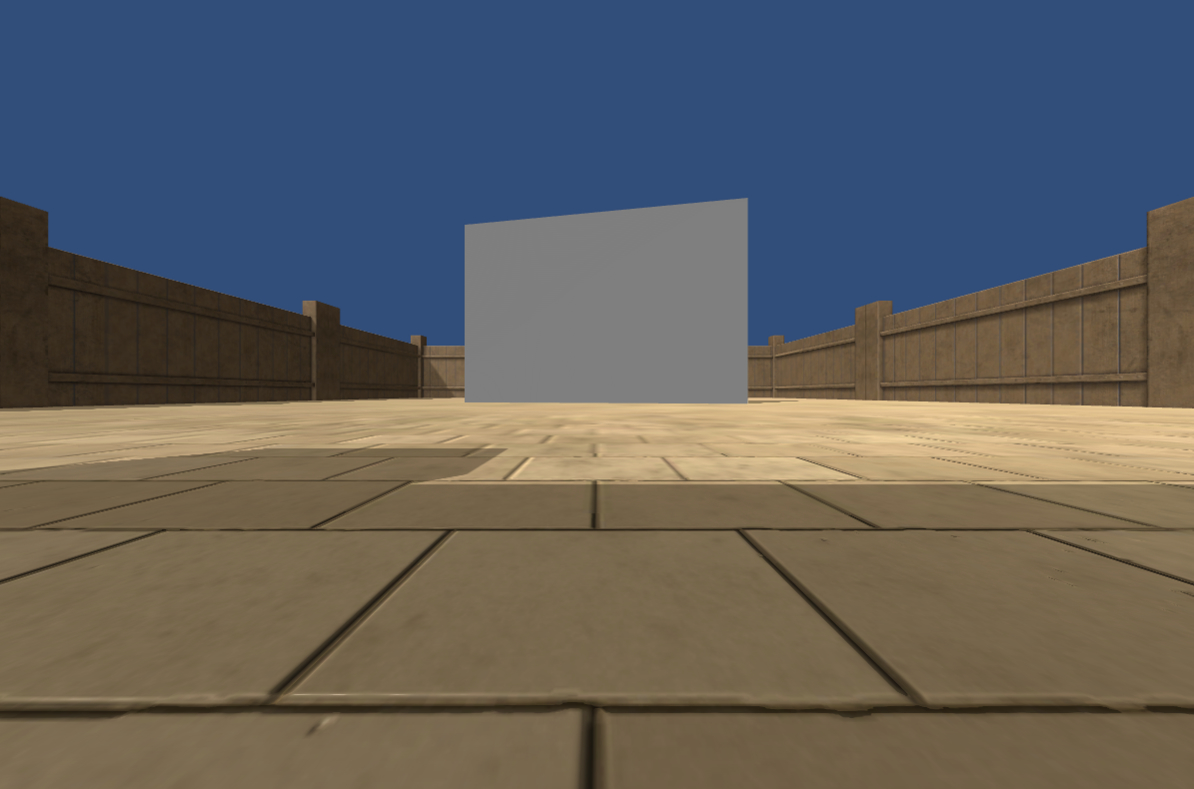}};
\node (im3) [xshift=-76ex,yshift=0ex]{\includegraphics[scale=0.075]{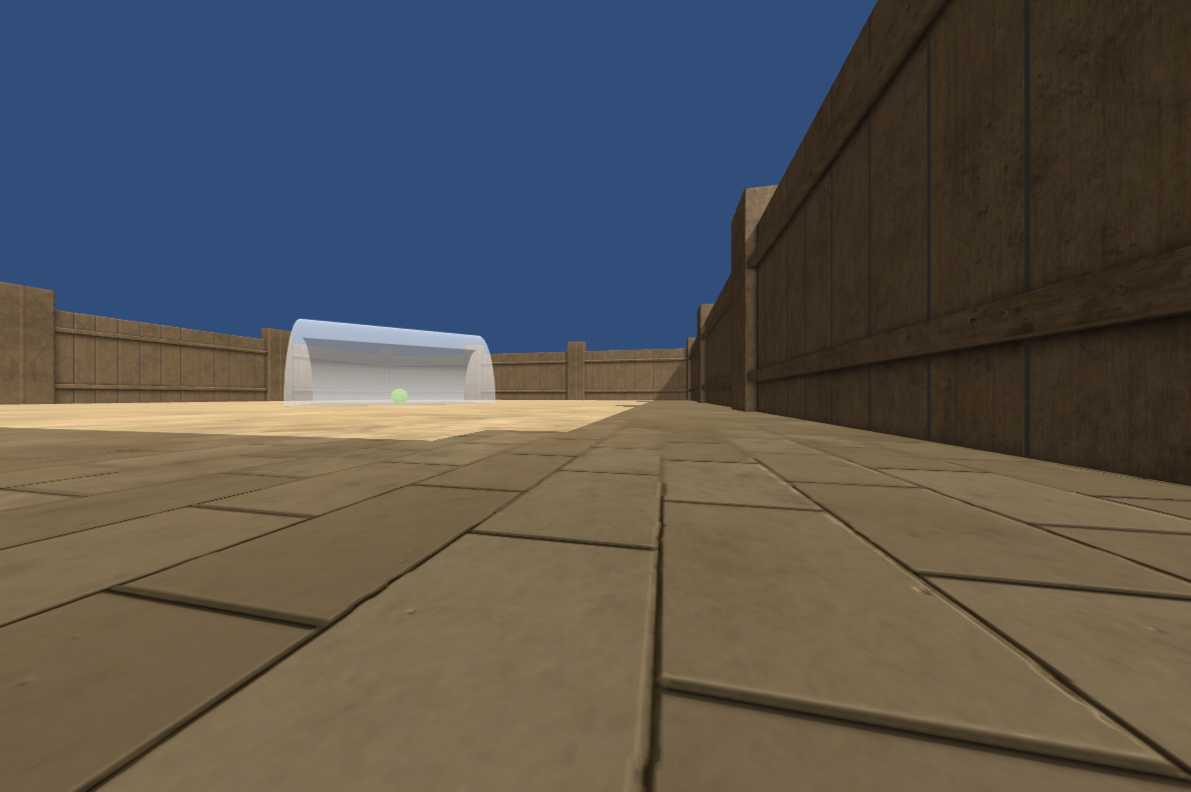}};
\node (im4) [xshift=-54ex, yshift=0ex]{\includegraphics[scale=0.075]{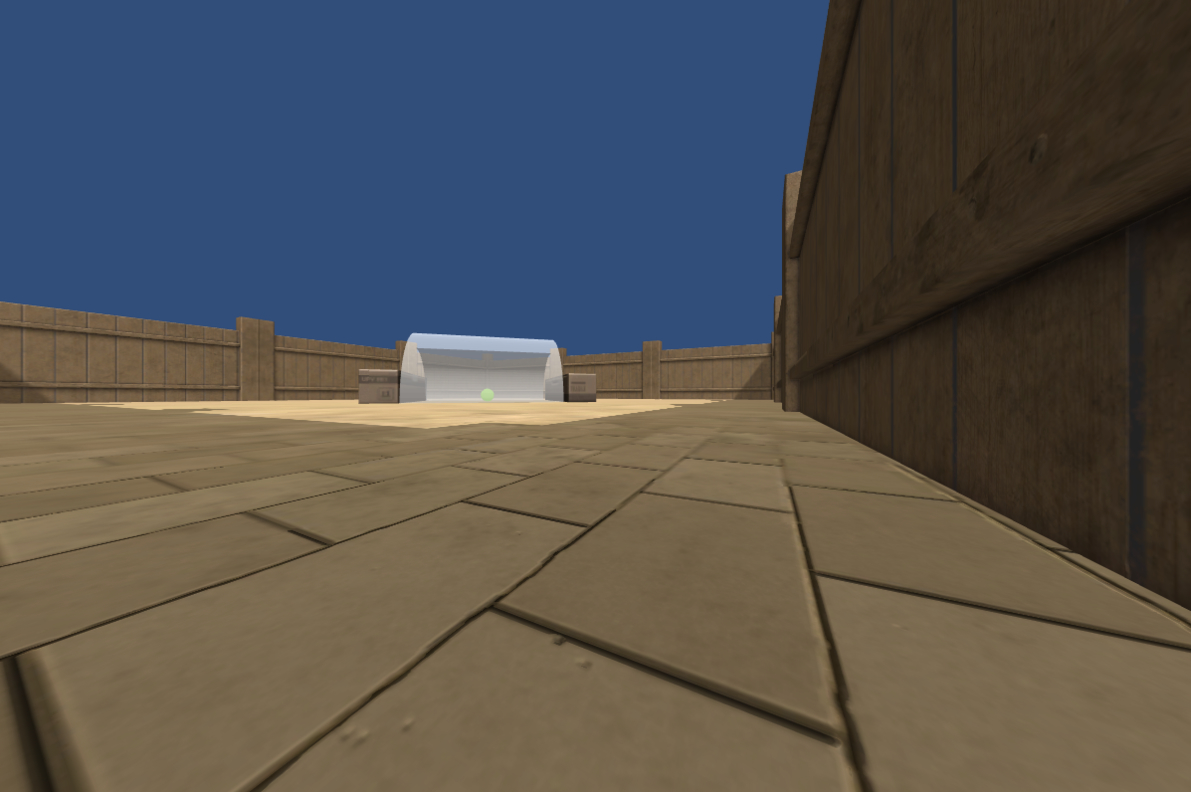}};
\node (im5) [xshift=-32ex, yshift=-0ex]{\includegraphics[scale=0.075]{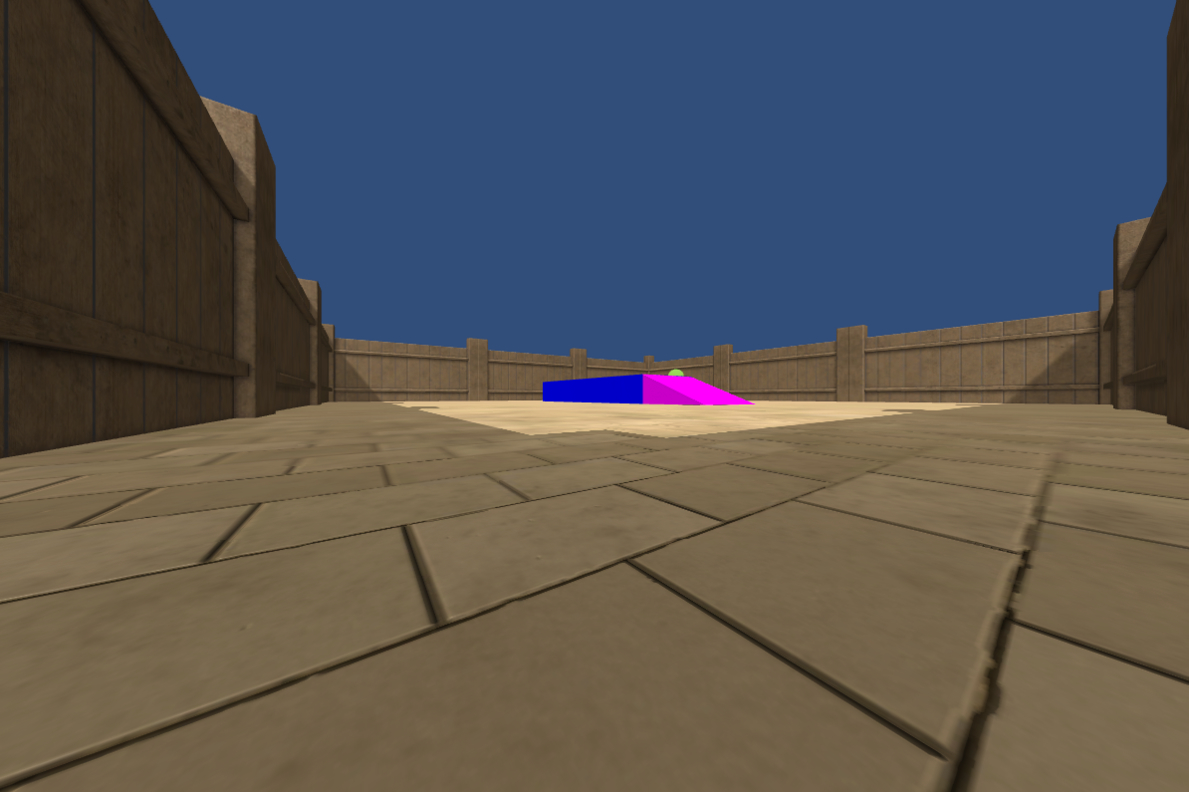}};
\node (a) [below of=im1, yshift=-2ex]{\footnotesize (a) Goal};
\node (b) [below of=im2, yshift=-2ex]{\footnotesize (b) Goal-behind wall};
\node (c) [below of=im3, yshift=-2ex]{\footnotesize (c) Goal-tunnel};
\node (d) [below of=im4, yshift=-2ex]{\footnotesize (d) Goal-occluded tunnel};
\node (e) [below of=im5, yshift=-2ex]{\footnotesize (e) Goal-on wall};
\end{tikzpicture}
\caption{All tasks feature one goal and one agent. The agent's and goal's positions are randomly selected at the start of each episode from a predefined set of fixed initial positions. Each episode initializes the environment by sampling these positions, ensuring variability while maintaining a structured distribution. There is only one source of reward per environment, i.e., a binary reward is provided for reaching the goal. The agent observes the arena through a first-person view with partial visibility, reflecting the limitations of a partially observable environment.}
\label{fig:arena}
\end{figure*}
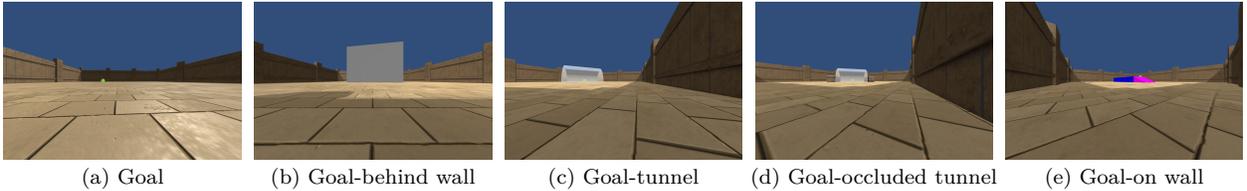
\subsubsection{Task Definitions:} The PointMaze environment is a 2-dimensional maze. We use two variants of the PointMaze environment. First, with fixed agent position and varying goal position, i.e., the goal position is reinitialized at each episode. Second, both the goal and agent positions are reinitialized after every reset. 

The Animal-AI Olympics~\citep{crosby2019animal} is a partially observable 3D environment in which an agent can navigate freely within an arena. We designed 5 experiments in total. Each experiment
has a different level of complexity based on the type of obstacles present in the arena. The descriptions of playgrounds are as follows :
\begin{figure*}[t]
\centering
\subfloat{\label{fig:im1}}
\subfloat{\label{fig:im2}}
\subfloat{\label{fig:im3}}
\subfloat{\label{fig:im4}}
\subfloat{\label{fig:im5}}
\begin{tikzpicture}
\node (im11) [xshift =-83ex,yshift = 8ex]{\includegraphics[scale=0.22]{leged.png}};
\node (im1) [xshift=-125ex, yshift = -5ex]{\includegraphics[scale=0.25]{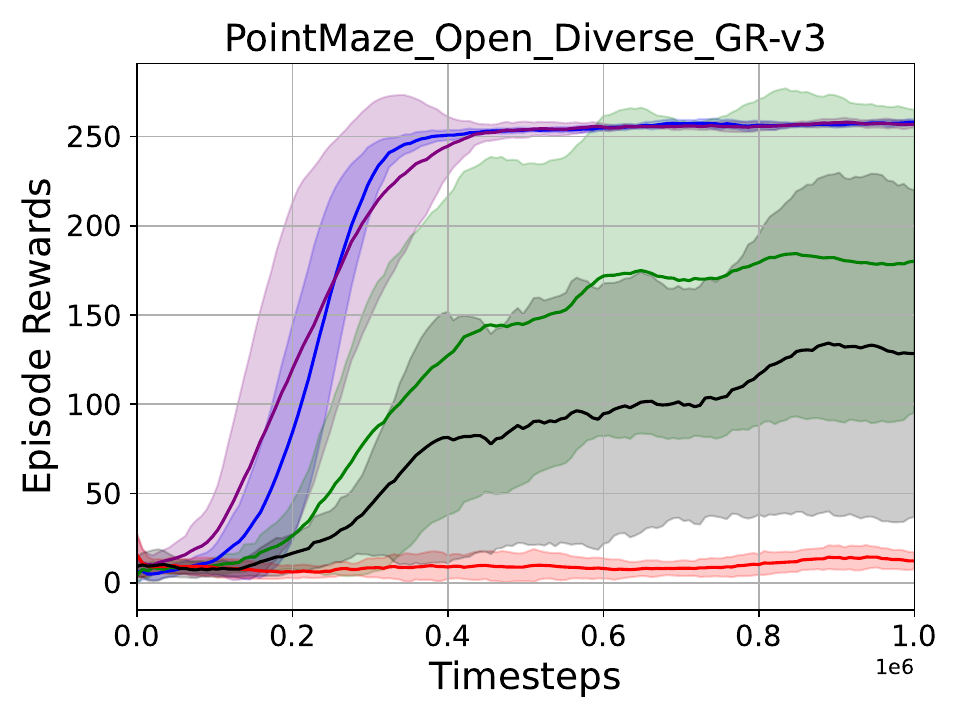}};
\node (im2) [xshift=-98.5ex,yshift = -5ex]{\includegraphics[scale=0.25]{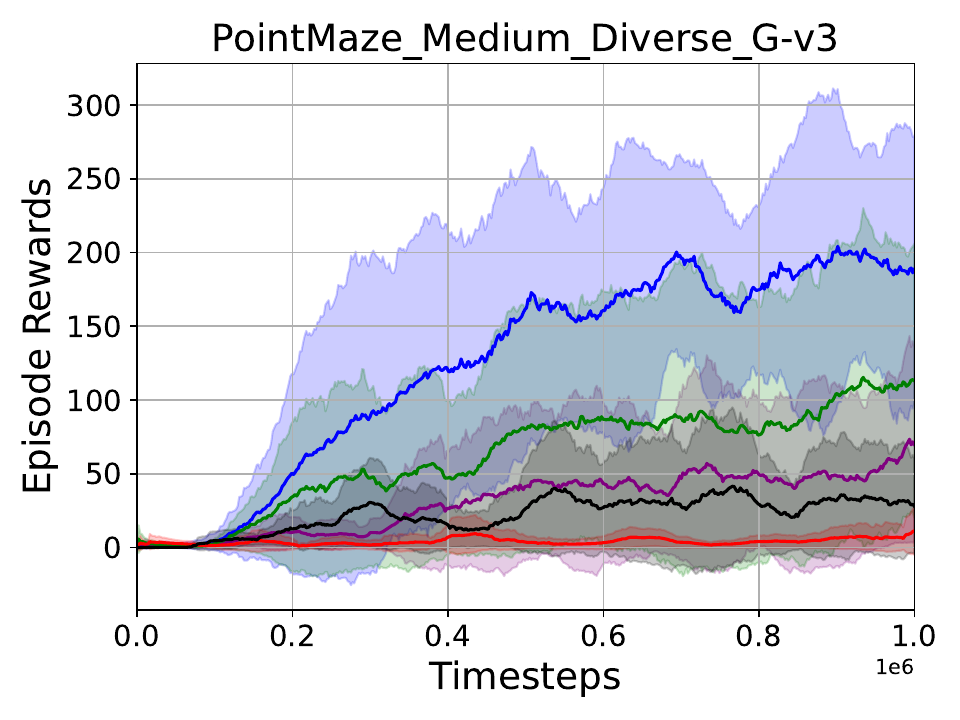}};
\node (im3) [xshift=-70.5ex,yshift = -5ex]{\includegraphics[scale=0.25]{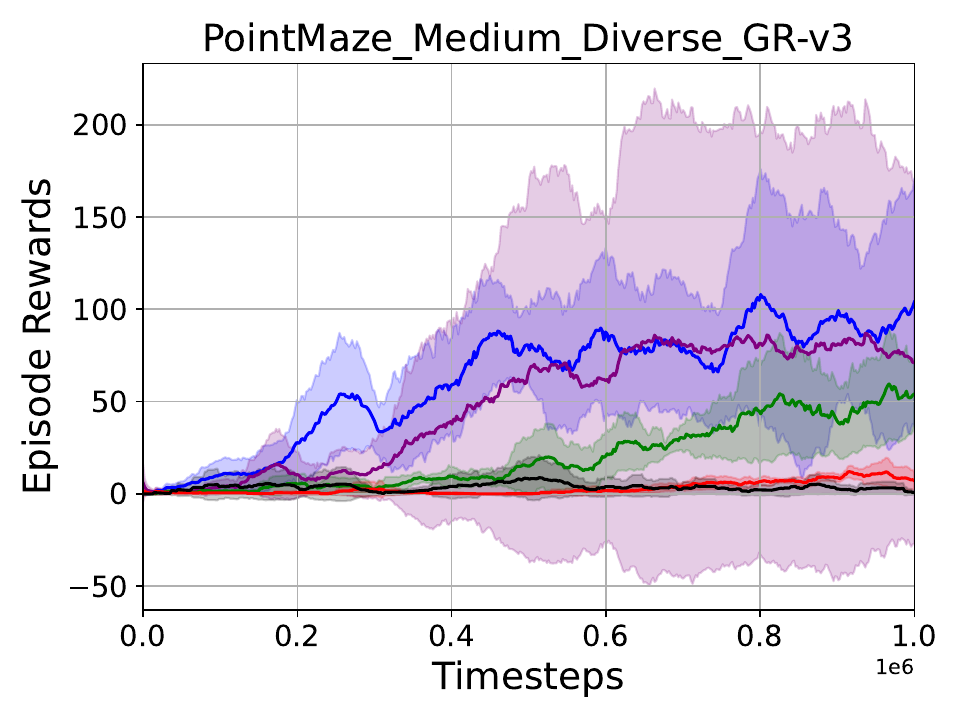}};
\node (im4) [xshift=-42.5ex, yshift = -5ex]{\includegraphics[scale=0.25]{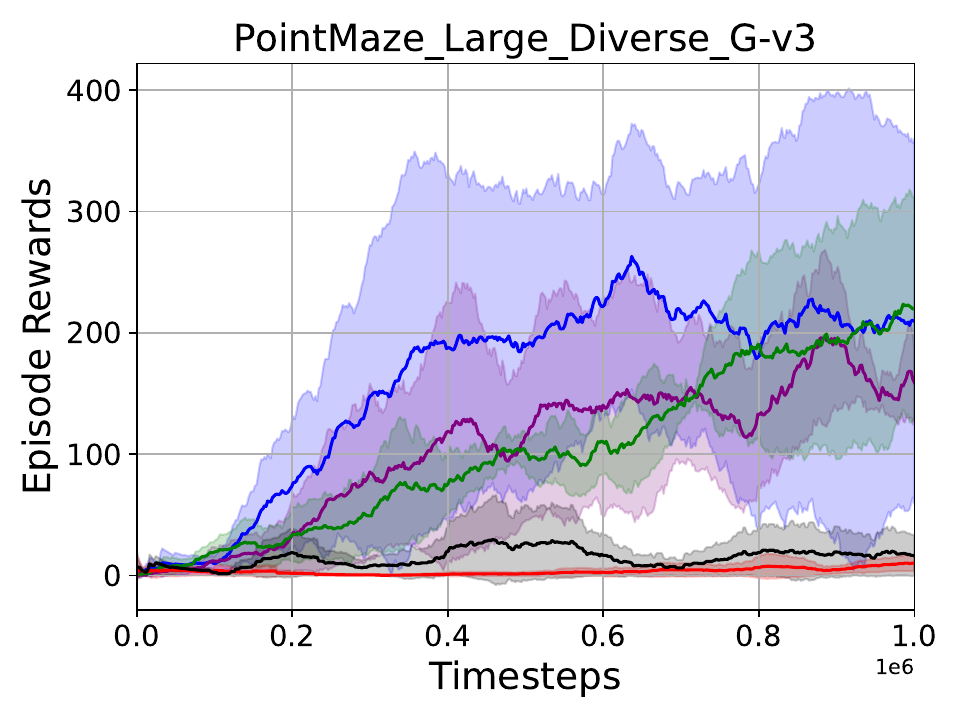}};

\end{tikzpicture}
\caption{Results show the performance on 4 PointMaze~\citep{gymnasium_robotics2023github} multi-goal sparse reward tasks (refer to Figure~\ref{fig:maze_result} for results on all tasks). The plots show the learning curves and the episodic rewards along the y-axis, evaluated under the current policy. The reported results are the mean across seven different seeds. The proposed algorithms outperform all the baselines by a significant margin.}
\label{fig:maze_result-1}
\end{figure*}

\begin{figure*}[t]
\centering
\subfloat{\label{fig:im1}}
\subfloat{\label{fig:im2}}
\subfloat{\label{fig:im3}}
\subfloat{\label{fig:im4}}
\subfloat{\label{fig:im5}}
\begin{tikzpicture}
\node (im11) [xshift =-83ex,yshift = 8ex]{\includegraphics[scale=0.22]{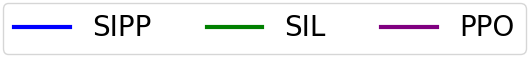}};
\node (im1) [xshift=-125ex, yshift = -5ex]{\includegraphics[scale=0.25]{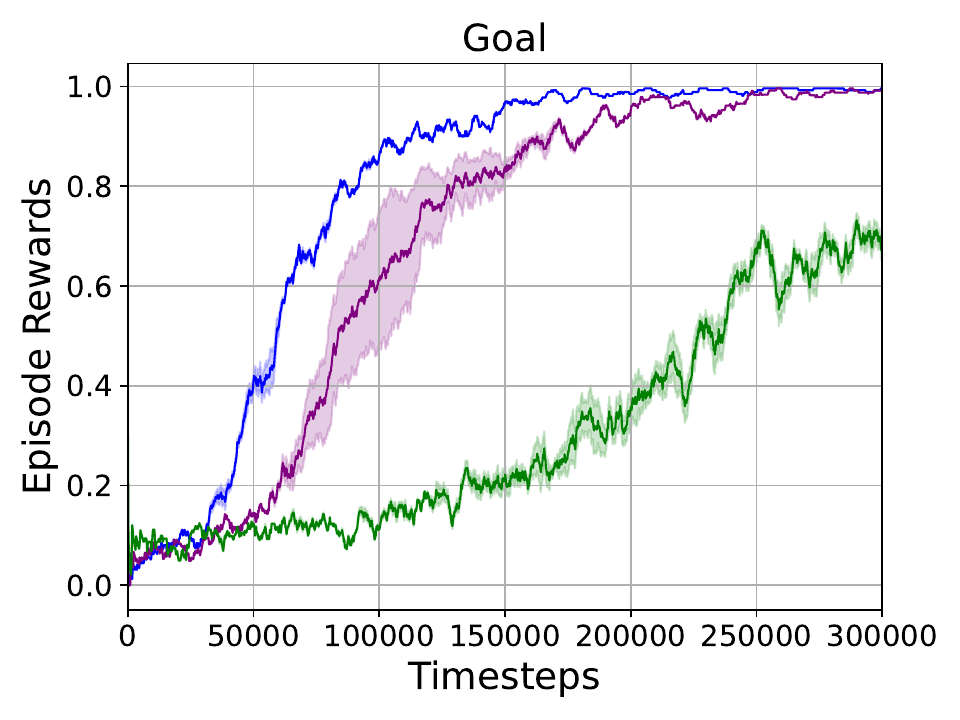}};
\node (im2) [xshift=-98.5ex,yshift = -5ex]{\includegraphics[scale=0.25]{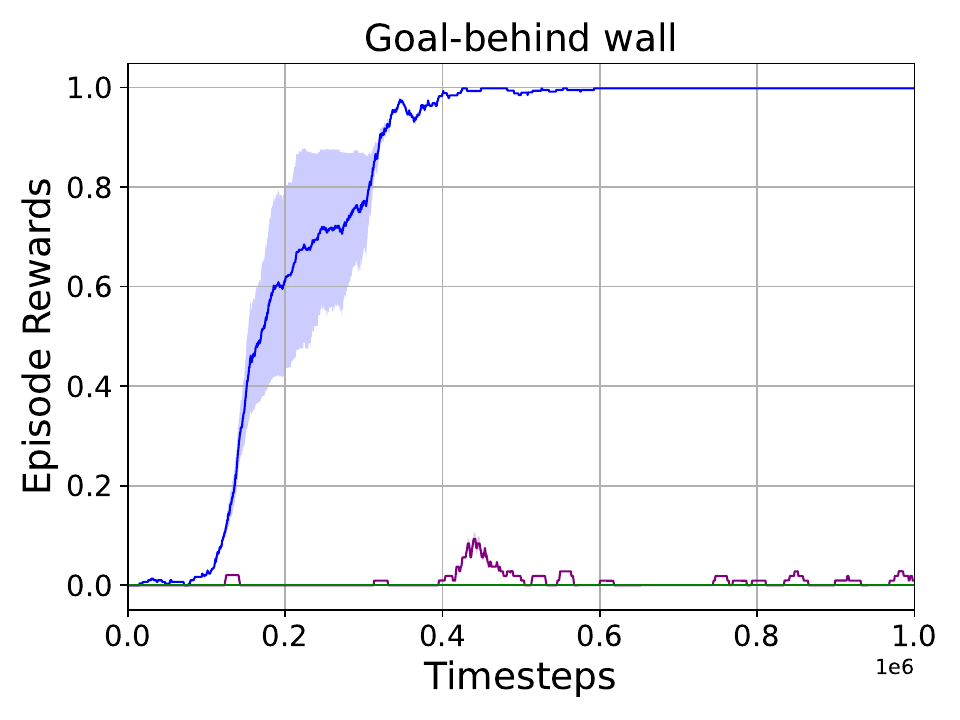}};
\node (im3) [xshift=-70.5ex,yshift = -5ex]{\includegraphics[scale=0.25]{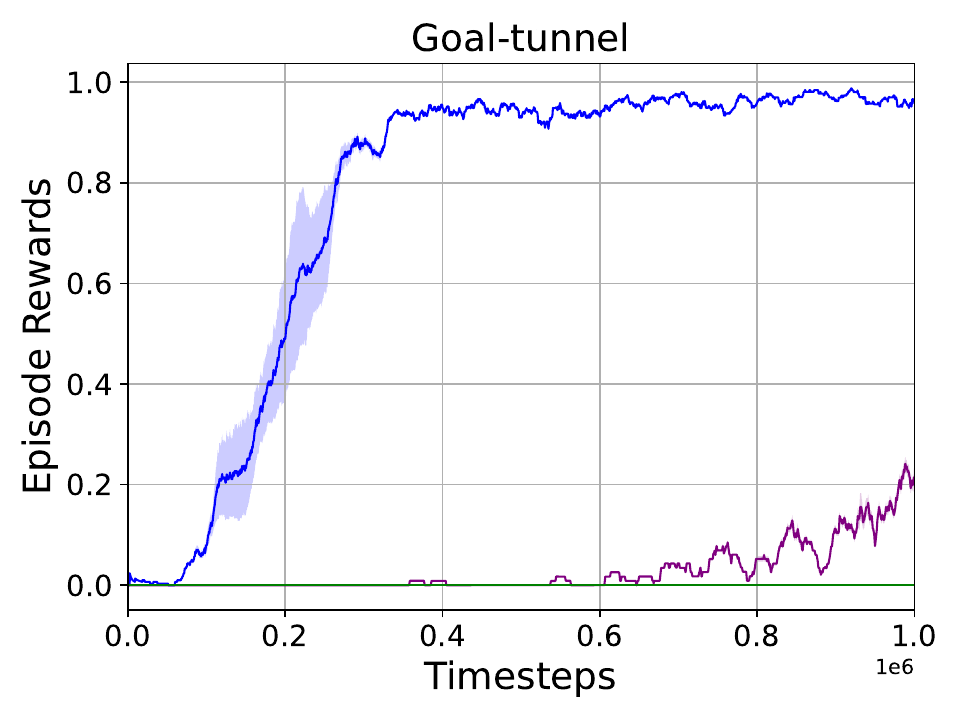}};
\node (im4) [xshift=-42.5ex, yshift = -5ex]{\includegraphics[scale=0.25]{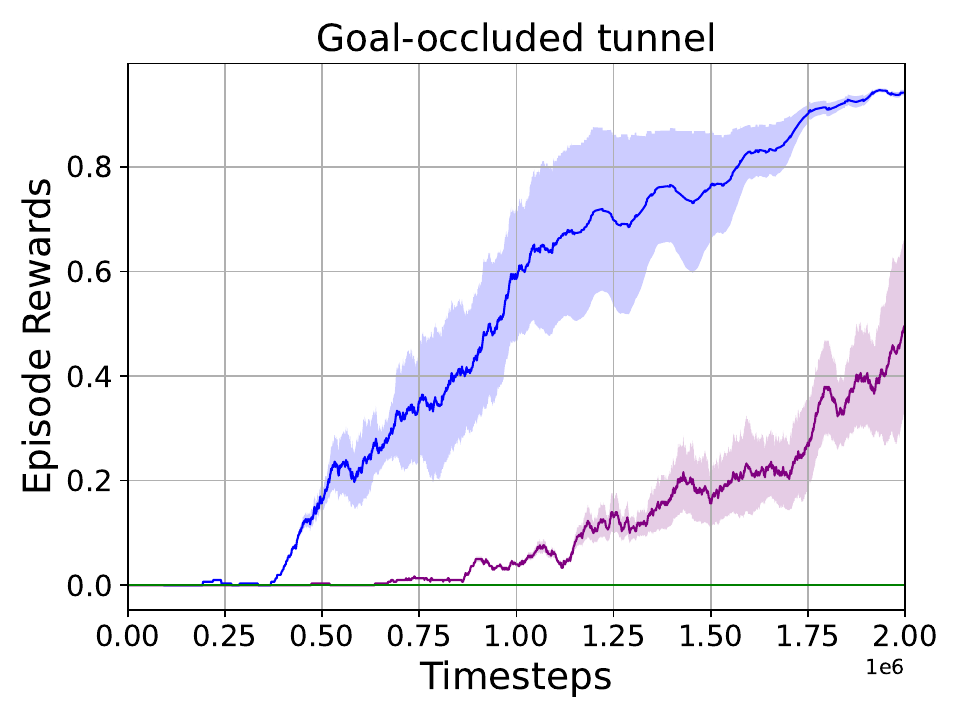}};

\end{tikzpicture}
\caption{Results show the performance on the 4 Animal-AI Olympics environment 
~\citep{crosby2019animal} binary reward tasks (refer to Figure~\ref{fig:animal} for results on all tasks). The plots show the learning curves, with episodic rewards (success rate) on the y-axis, evaluated under the current policy. The reported results are the mean across 5 seeds, with shaded regions highlighting the standard deviation. The proposed algorithms outperform PPO by a significant margin.}
\label{fig:animal-1}
\end{figure*}
\begin{itemize}
    \item \textbf{Goal:}    In this arena, the agent has to reach the goal position. The agent and goal can be anywhere in the arena. There are no obstacles in the arena.
    \item \textbf{Goal-behind wall}:    
    The goal is hidden behind a wall in this arena. The agent and goal positions are different in each configuration. The agent needs to learn to find the goal, which is hidden behind the wall.
    \item \textbf{Goal-tunnel}: 
    This arena has a transparent tunnel open at both ends. The agent cannot penetrate the tunnel walls and must enter the tunnel to reach the goal. 
    \item \textbf{Goal-occluded tunnel:}    
    This arena is identical to the previous one, except that the tunnel entrances are occluded with movable boxes. The agent must learn to move the boxes to find the goal inside the tunnel.
    \item \textbf{Goal-on wall:}    In this arena, we place the goal on an L-shaped wall. The agent must learn to find a ramp to climb up the wall and avoid falling off the wall to reach the goal.
\end{itemize}

\subsubsection{Empirical Analysis:}
\begin{figure*}[t]
\centering
\subfloat{\label{fig:im11}}
\subfloat{\label{fig:im1}}
\subfloat{\label{fig:im2}}
\subfloat{\label{fig:im3}}
\subfloat{\label{fig:im4}}
\begin{tikzpicture}
\node (im11) [xshift =-83ex,yshift = 8ex]{\includegraphics[scale=0.22]{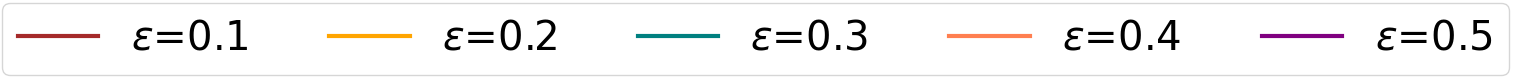}};
\node (im1) [xshift=-125ex, yshift = -5ex]{\includegraphics[scale=0.25]{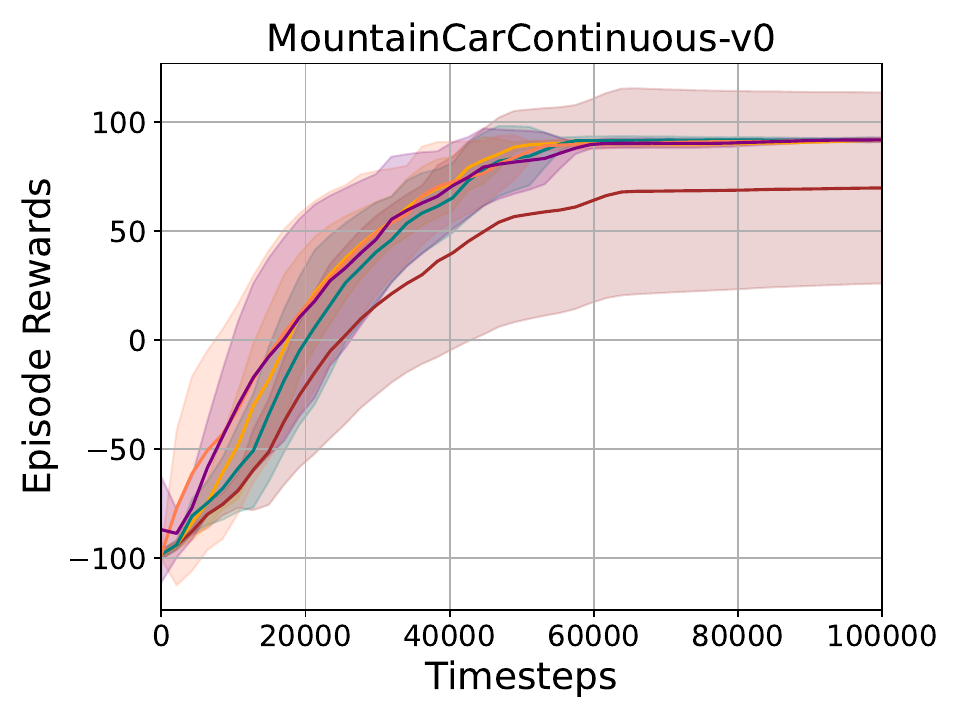}};
\node (im2) [xshift=-98.5ex,yshift = -5ex]{\includegraphics[scale=0.25]{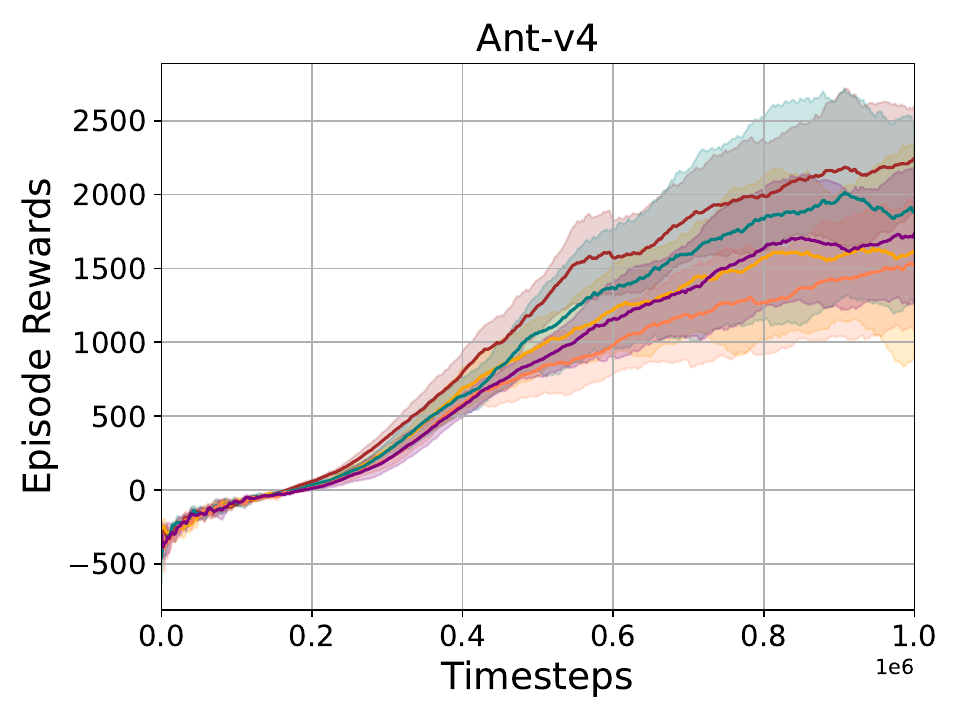}};
\node (im3) [xshift=-70.5ex,yshift = -5ex]{\includegraphics[scale=0.25]{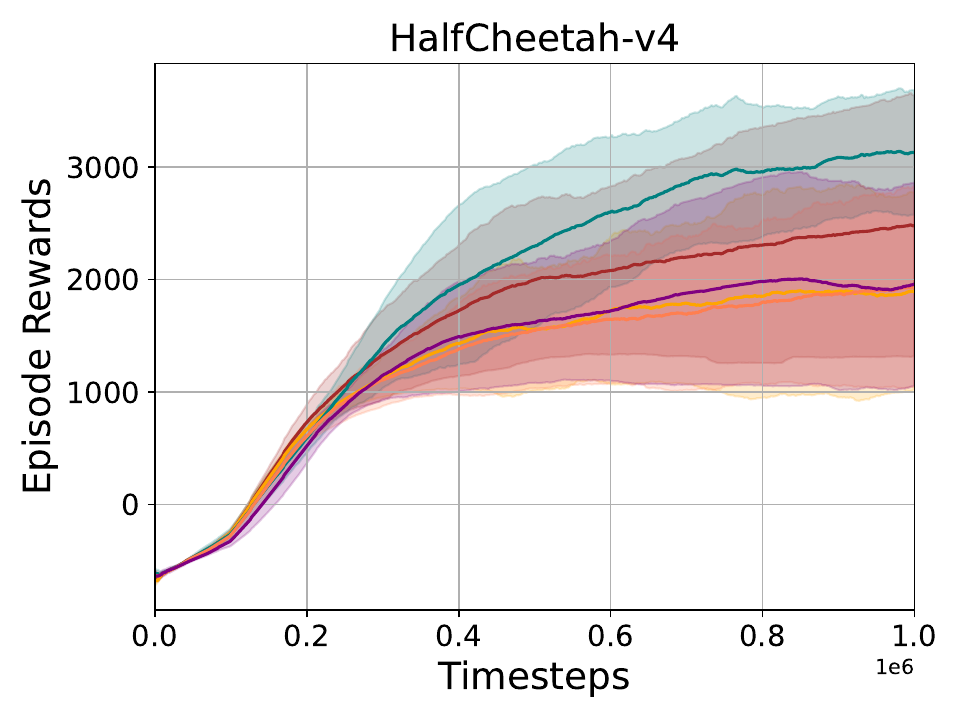}};
\node (im4) [xshift=-42.5ex, yshift = -5ex]{\includegraphics[scale=0.25]{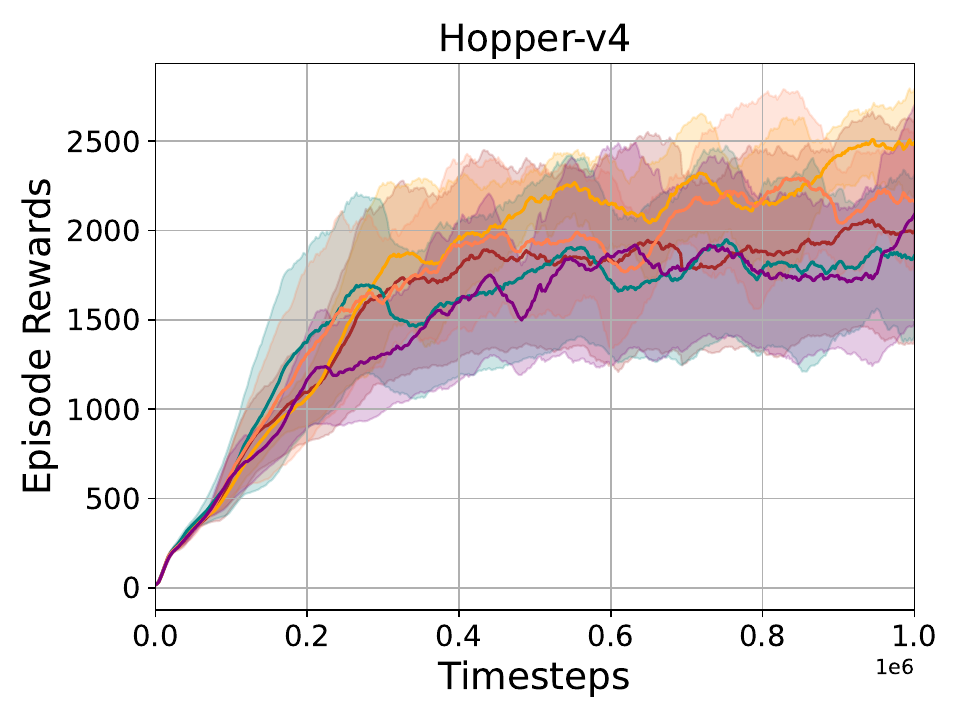}};

\end{tikzpicture}
\caption{Results show the ablation study on 4 MuJoCo~\citep{towers_gymnasium_2023} continuous control tasks (refer to Figure~\ref{fig:match_abe} for complete results). The parameter $\epsilon$ controls the balance between exploration and exploitation. The plots show the learning curves and the episodic rewards along the y-axis, evaluated under the current policy with different $\epsilon$. The reported results are the mean across 5 seeds, with shaded regions highlighting the standard deviation.}

\label{fig:match_abe-1}
\end{figure*}

\begin{figure*}[t]
\centering
\subfloat{\label{fig:im1}}
\subfloat{\label{fig:im2}}
\subfloat{\label{fig:im3}}
\subfloat{\label{fig:im4}}
\subfloat{\label{fig:im5}}
\begin{tikzpicture}
\node (im11) [xshift =-83ex,yshift = 8ex]{\includegraphics[scale=0.22]{legend_abe.png}};
\node (im1) [xshift=-125ex, yshift = -5ex]{\includegraphics[scale=0.25]{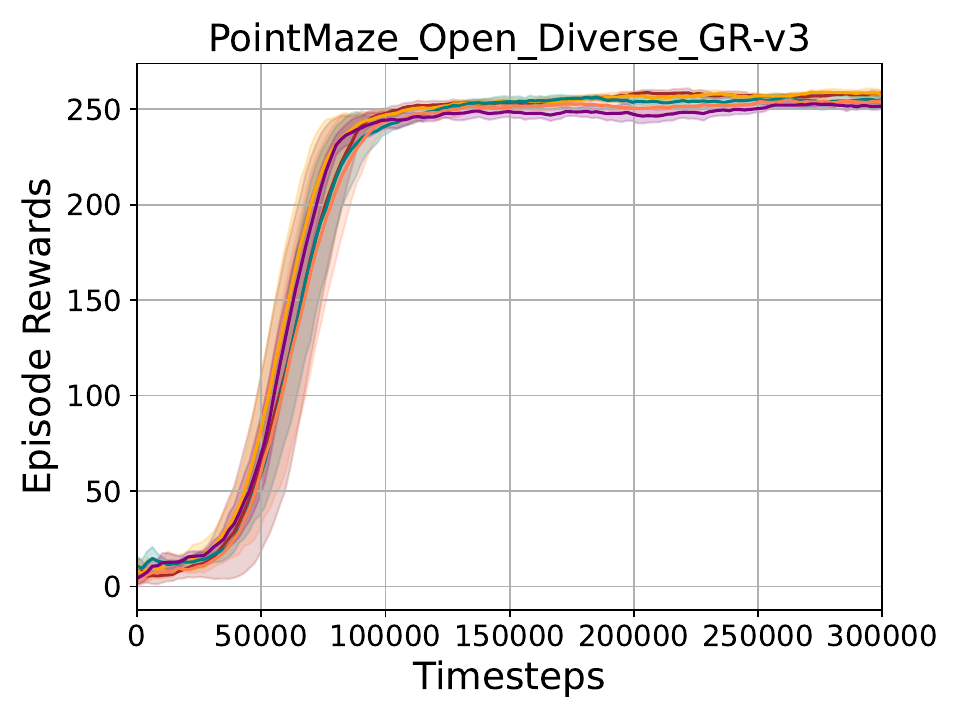}};
\node (im2) [xshift=-98.5ex,yshift = -5ex]{\includegraphics[scale=0.25]{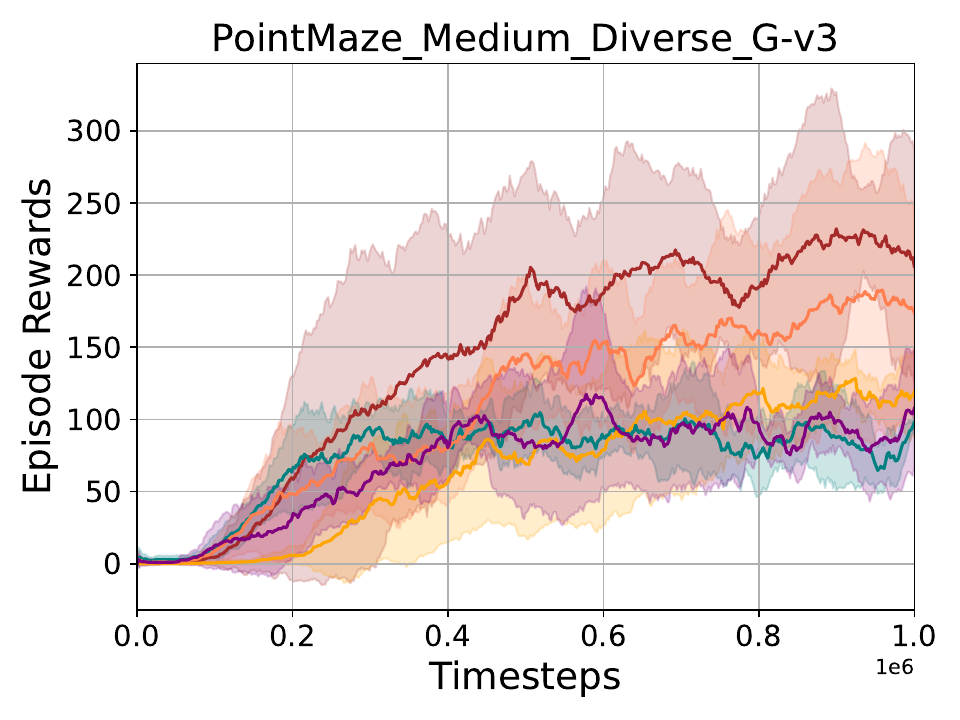}};
\node (im3) [xshift=-70.5ex,yshift = -5ex]{\includegraphics[scale=0.25]{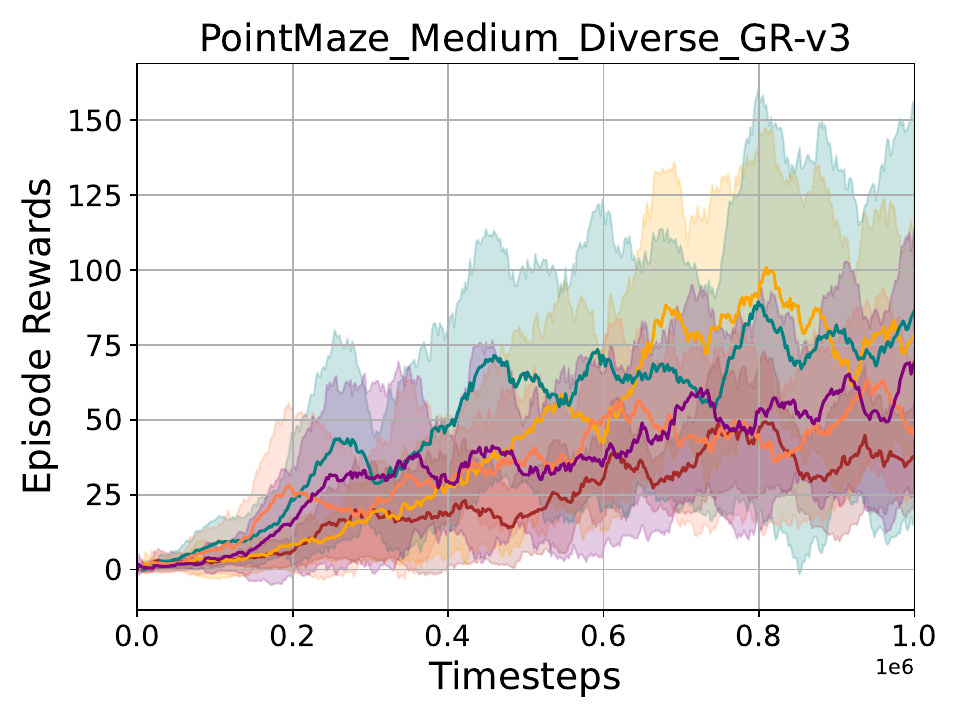}};
\node (im4) [xshift=-42.5ex, yshift = -5ex]{\includegraphics[scale=0.25]{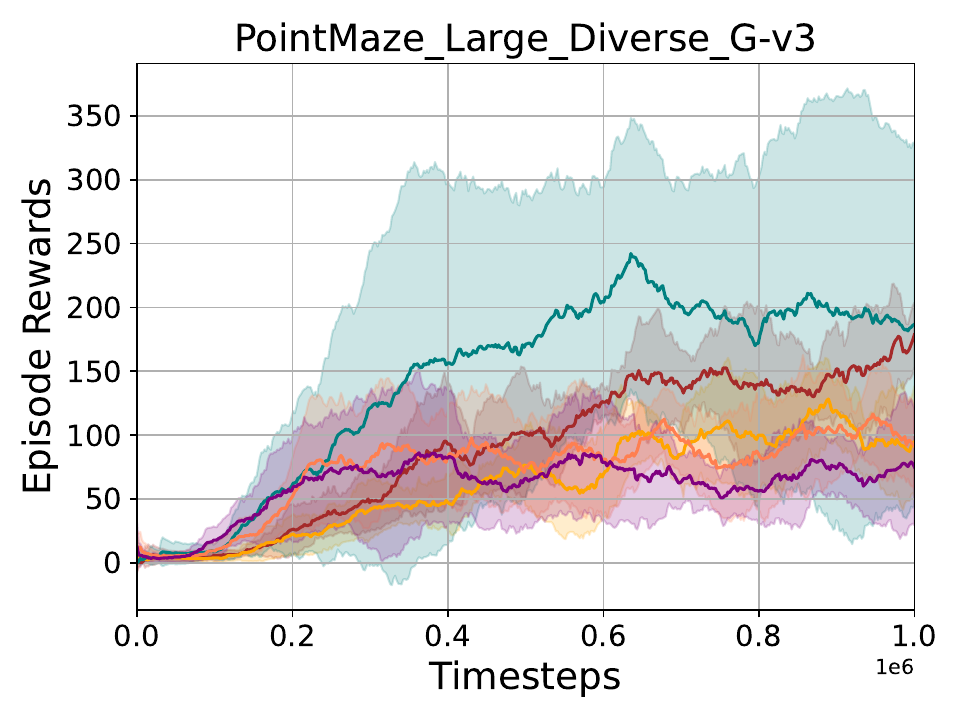}};

\end{tikzpicture}
\caption{Results show the ablation study on 4 PointMaze~\citep{gymnasium_robotics2023github} multi-goal sparse reward tasks (refer to Figure~\ref{fig:maze_abe} for complete results). The parameter $\epsilon$ controls the replay frequency to balance exploration vs exploitation. The plots show the learning curves and the episodic rewards along the y-axis, evaluated under the current policy with different $\epsilon$. The reported results are across 5 different seeds.}
\label{fig:maze_abe-1}
\end{figure*}
The performance of \textit{SIPP-Replay} is shown in Figures~\ref{fig:maze_result-1} and \ref{fig:animal-1}. The choice of baselines for the PointMaze environment is consistent with the MuJoCo tasks, as both involve fully observable MDPs. However, for the Animal-AI Olympics task, the choice of baselines is restricted to PPO. This limitation arises because the official implementations of baselines, SIL, and SVPG are tailored to fully observable environments like MuJoCo and do not support the Animal-AI Olympics. 

Further, Baselines such as SIL and SVPG rely on explicit divergence estimation or advantage comparisons over fully observed states, making them less suitable for partially observable environments. In contrast, SIPP’s replay strategy treats stored trajectories as expert demonstrations and integrates them directly into PPO’s training without requiring density estimation or changes to the underlying architecture.
This allows SIPP to operate naturally in POMDP settings, where modeling state visitation distributions is non-trivial or infeasible.

To ensure fairness, we adapted the SIL baseline~\citep{oh2018self} by using observation histories as proxies for states in the Animal-AI Olympics environment. However, even with this adaptation, SIL performed worse than vanilla PPO. This degradation stems from SIL’s reliance on either dense reward signals or access to true states to guide its imitation strategy—conditions that are not satisfied in our partially observable setting.
This comparison highlights the broader applicability of our method: SIPP does not assume full observability or reward density and is, to our knowledge, the first self-imitation approach successfully deployed in such a complex, diverse POMDP environment.

In our experiments, the Imitation-Exploration Trade-off coefficient IET ($\epsilon$) was set to 0.3 for all PointMaze tasks except for PointMaze\_Medium\_Diverse\_G, where $\epsilon=0.1$ was used based on preliminary experiments. The performance of our proposed algorithm exceeds all the baselines on the Maze navigation task, as shown in Figure~\ref{fig:maze_result-1}. We believe that the long-term dependency problem plays a crucial role under sparse reward conditions, and we tackle this by replaying past trajectories. Unlike baseline methods, which prioritize state-action pairs, SIPP focuses on episodic trajectory-level prioritization rather than state visitation distributions. This helps the agent to understand which actions contribute to future rewards. To incorporate successful trajectories in the replay buffer, we consider it as the possible behavior of the agent in the current environment. 

The \textit{Replay} strategy results in even superior performance (Figure~\ref {fig:animal-1}) in partially observable environments due to its inherent ability to adapt to partial observability. The results show that PPO agents encounter some success but fail to learn due to policy instability. Policy instability refers to the divergence of PPO’s policy from successful behaviors due to frequent updates with new data, which overwrite past successes. However, the proposed~\textit{Replay} strategy stores these behaviors and repeatedly replays them. This reinforces successful behaviors in policy learning, and the agent eventually learns to mimic them.

To summarize, the results show that self-imitation can help agents learn in both dense and sparse reward settings. In a dense reward setting, prioritizing state visitation that matches the past successful state is sufficient, as a dense reward structure can guide the agent to learn the long-term consequences of actions taken in those states. However, a more exploitative strategy is required in sparse reward settings, such as replaying past successful episodic trajectories. 

\subsection{Tuning Self-Imitation vs. Exploration}
The proposed strategy exploits the agent's past behaviors. However, it is crucial for the agent to learn an optimal policy. This balance between exploitation and exploration in SIPP is achieved through the Imitation-Exploration Trade-off (IET) coefficient $\epsilon$. In \textit{SIPP-Match} this parameter indicates the probability of sampling training batches uniformly or with a priority proportional to the optimal transport distance with the most successful past trajectory. In \textit{SIPP-Replay}, this parameter controls the trajectory replay probability from the imitation buffer $\mathcal{B_I}$. 

The effect of IET for \textit{SIPP-Match} strategy is shown in Figure \ref{fig:match_abe-1}. The ablation study shows the maximum performance improvement for $\ epsilon=0.1, 0.2$, or $0.3$ across all tasks. This highlights the importance of exploration, as greedy imitation results in sub-optimal performance. However, for simpler tasks such as MountainCarContinuous, the performance is mostly similar as these simpler environments require less exploration, and even an aggressive imitation strategy results in similar performance.

A similar analysis was performed to find the balance between the exploration and exploitation trade-off of the PointMaze navigation environments. The results of this ablation study are shown in Figure \ref{fig:maze_abe-1}. We didn't perform a similar analysis of the Animal-AI Olympics environment. However, the IET coefficient for the Animal-AI Olympics environment was kept $\epsilon=0.3$ based on the insights from the above ablation studies. 

In conclusion, our ablation study on $\epsilon$ highlights its influence on SIPP’s performance. In dense reward environments like MuJoCo, smaller $\epsilon= 0.1,0.2$ perform well, as frequent rewards naturally guide exploration. Conversely, in sparse-reward settings like PointMaze, a higher value of $\epsilon=0.3$ improves outcomes by emphasizing imitation of rare successful trajectories. These results indicate that $\epsilon$ should be adjusted based on the task’s reward structure and exploration demands.

\section{Conclusion}
This paper proposes a self-imitating proximal policy framework to address exploration and sample-efficiency challenges in dense and sparse-reward environments. Through extensive experimentation, we demonstrated that bootstrapping policy learning from past rewarding experiences effectively reduces policy divergence, leading to enhanced exploration and stability. The simplicity and efficacy of the proposed algorithm highlight its versatility across different problem settings. Furthermore, we showed that self-imitation and exploration are inherently complementary, enabling agents to leverage prior successes for guided learning, which can be crucial in hard exploration tasks.

\clearpage

\bibliography{main}
\bibliographystyle{main}
\newpage
\appendix

\section{Integrating SIPP with Hard Exploration Techniques}
In this section, we investigate integrating the Self-Imitating Proximal Policy (SIPP) with Random Network Distillation (RND), a technique that enhances exploration in reinforcement learning by encouraging agents to visit novel states. This combination leverages SIPP's self-imitation mechanism, which reinforces past successful behaviors, and RND's intrinsic motivation, which promotes exploration, to improve performance in environments with challenging exploration requirements.

Our approach combines two core components:
\begin{itemize}
    \item Self-Imitating Proximal Policy (SIPP): SIPP enhances policy optimization by prioritizing high-return trajectories from the agent's past experiences. In sparse reward settings, such as Atari games, we employ the Replay strategy, maintaining an imitation buffer of successful trajectories (defined by cumulative extrinsic rewards) that are selectively replayed during policy updates using Proximal Policy Optimization (PPO).
    \item Random Network Distillation (RND): RND generates an intrinsic reward signal based on the prediction error between a fixed random network and a trainable network, incentivizing the agent to explore novel states by quantifying their unfamiliarity.
\end{itemize}

In the RND+SIPP framework, the agent's total reward at each timestep is the sum of the extrinsic reward from the environment and the intrinsic reward from RND. The policy is updated via PPO with the SIPP Replay strategy, where the imitation buffer stores trajectories based solely on their cumulative extrinsic rewards. This ensures that SIPP reinforces behaviors that lead to tangible environmental success, while RND independently explores novel regions, mitigating potential conflicts between exploitation and exploration objectives.

We evaluated RND+SIPP on three Atari 2600 games from the Arcade Learning Environment—Gravitar, Venture, and Solaris—selected for their sparse rewards and exploration challenges. The results of RND were taken from~\citep{burda2018exploration}. We did not perform an ablation study on the SIPP hyperparameter for this study and kept the IET coefficient fixed across tasks to 0.1 and the imitation buffer size 1. In the Gravitar task, RND+SIPP achieves an 11.7\% improvement, leveraging SIPP’s reinforcement of successful trajectories alongside RND’s exploration. Venture Performance is comparable, with a slight 2.5\% decrease, suggesting task-specific tuning of \(\epsilon\) may be needed. In Solaris, A 9.3\% gain highlights the benefit of combining imitation with exploration in complex state spaces.

\begin{table}[h]
    \centering
    \caption{Performance Comparison on hard exploration tasks.}
    \label{tab:performance_comparison}
    \begin{tabular}{lcc}
        \toprule
        \textbf{Task} & \textbf{RND} & \textbf{RND+SIPP} \\
        \midrule
        Gravitar  & 3906  & \textbf{4363}  \\
        Venture   & \textbf{1859}  & 1813  \\
        Solaris   &3282  & \textbf{3589} \\
        \bottomrule
    \end{tabular}
\end{table}

The integration of SIPP with RND demonstrates that combining self-imitation learning with intrinsic motivation provides a dual benefit. On the one hand, SIPP ensures that the agent leverages its past successes to stabilize policy updates. On the other hand, RND continually drives the agent to explore unvisited or less familiar regions of the state space. The trade-off between these components is controlled by the Imitation-Exploration Trade-off coefficient, enabling task-specific tuning.

The analysis suggests that while RND alone can foster exploration, it may not prevent the divergence of effective strategies. RND+SIPP overcomes this limitation by continually reinforcing high-value behaviors, thereby improving overall performance. Future work may involve dynamically adapting the balance between intrinsic and extrinsic rewards based on the observed learning dynamics, thereby further refining the exploration-exploitation trade-off.

\section{Guidelines for Tuning \(\epsilon\)-Greedy Parameter in SIPP}
The hyperparameter \(\epsilon\) governs the balance between imitation and exploration in SIPP. To tune it effectively, we suggest the following:
\begin{itemize}
  \item \textbf{Baseline Setting:} Start with \(\epsilon = 0.3\), which works reasonably well across diverse tasks.
  \item \textbf{Dense Reward Tasks:} For environments with frequent rewards (e.g., MuJoCo), reduce \(\epsilon\) to 0.1–0.2 to prioritize refining known behaviors over excessive exploration.
  \item \textbf{Sparse Reward Tasks:} In hard exploration scenarios (e.g., PointMaze), increase \(\epsilon\) to 0.3–0.5 to leverage imitation of scarce successes.
  \item \textbf{Stability Check:} If training shows a high variance in success rates, consider reducing \(\epsilon\) to stabilize learning via stronger imitation.
  \item \textbf{Cross-Validation:} For optimal results, perform a grid search over values like 0.1, 0.3, and 0.5, especially in critical applications.
\end{itemize}
These guidelines, grounded in our empirical findings, should enhance SIPP’s reproducibility and usability.

\begin{figure*}[tbh]
\centering
\subfloat{\label{fig:im11}}
\subfloat{\label{fig:im1}}
\subfloat{\label{fig:im2}}
\subfloat{\label{fig:im3}}
\subfloat{\label{fig:im4}}
\subfloat{\label{fig:im6}}
\subfloat{\label{fig:im7}}
\subfloat{\label{fig:im9}}
\subfloat{\label{fig:im10}}
\begin{tikzpicture}
\node (im11) [xshift =-83ex,yshift = 8ex]{\includegraphics[scale=0.22]{leged.png}};
\node (im1) [xshift=-125ex, yshift = -5ex]{\includegraphics[scale=0.25]{MountainCarContinuous-v0.pdf}};
\node (im2) [xshift=-98.5ex,yshift = -5ex]{\includegraphics[scale=0.25]{Ant-v4.pdf}};
\node (im3) [xshift=-70.5ex,yshift = -5ex]{\includegraphics[scale=0.25]{HalfCheetah-v4.pdf}};
\node (im4) [xshift=-42.5ex, yshift = -5ex]{\includegraphics[scale=0.25]{Hopper-v4.pdf}};
\node (im6) [xshift=-125ex, yshift=-28ex]{\includegraphics[scale=0.25]{InvertedDoublePendulum-v4.pdf}};
\node (im7) [xshift=-98.5ex, yshift=-28ex]{\includegraphics[scale=0.25]{InvertedPendulum-v4.pdf}};
\node (im9) [xshift=-70.5ex, yshift=-28ex]{\includegraphics[scale=0.25]{HumanoidStandup-v4.pdf}};
\node (im10) [xshift=-42.5ex, yshift=-28ex]{\includegraphics[scale=0.25]{Walker2d-v4.pdf}};
\node (im5) [xshift=-98.5ex, yshift=-51ex]{\includegraphics[scale=0.25]{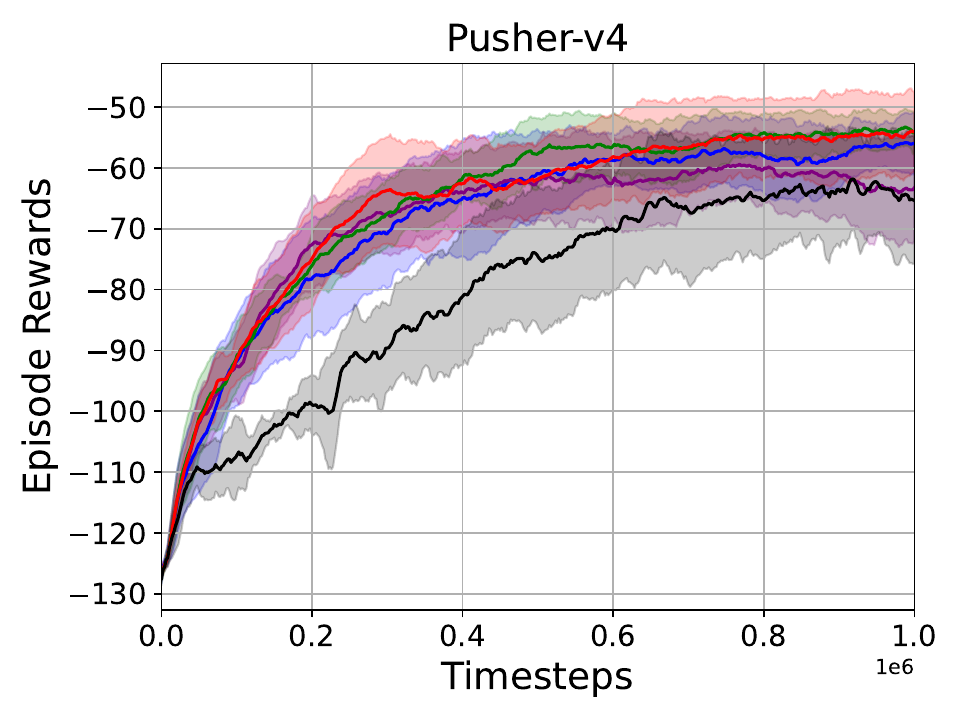}};
\node (im8) [xshift=-70.5ex, yshift= -51ex]{\includegraphics[scale=0.25]{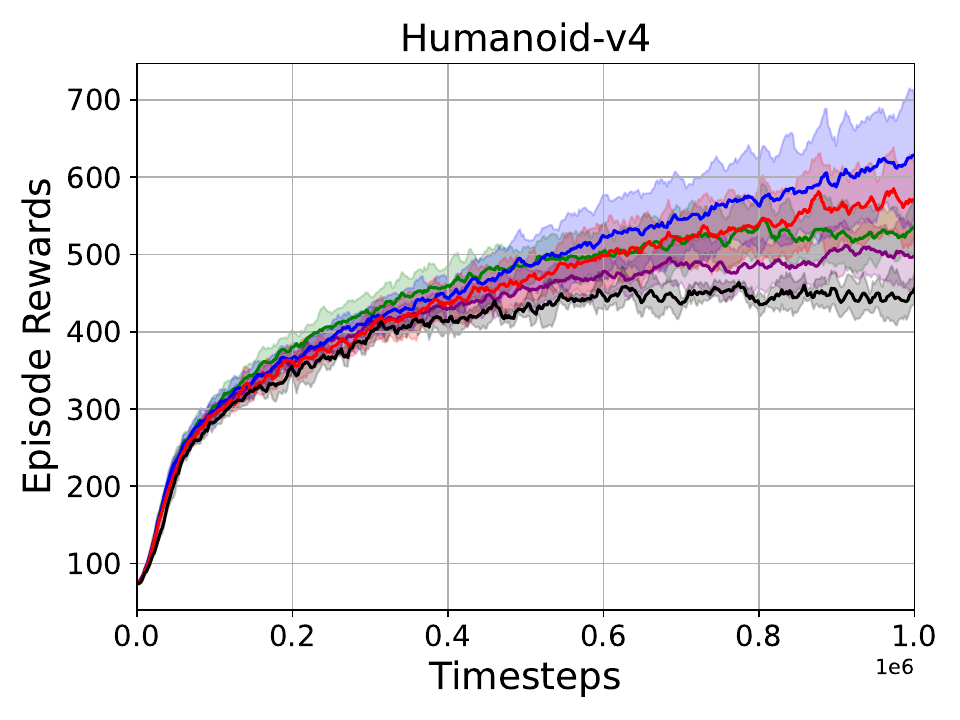}};

\end{tikzpicture}
\caption{Results show the performance of 10 MuJoCo~\citep{towers_gymnasium_2023} continuous control tasks. The plots are the learning curves and show the episodic rewards along the y-axis, evaluated through the current policy. The reported results are the mean across seven different seeds. The proposed algorithms outperform all baselines across all tasks by being competitive or better than others. }

\label{fig:SIPP-match}
\end{figure*}

\begin{table}[h!]
\centering
\begin{minipage}{0.45\textwidth}
    \centering
    \caption{Parameters For Animal-AI Olympics Environment}
    \label{table1}
    \setlength{\tabcolsep}{16pt}
    \begin{tabular}{lcr}
        \toprule
        Parameter  & Values \\
        \midrule
        episode length  & 1000 \\
        image size (RGB) & $84 \times 84 \times 3$ \\
        initial reward threshold & 0\\
        frame-skip     & 2 \\
        frame-stack       & 4 \\
        discount factor ($\lambda$) & 0.99\\
        gae-gamma ($\gamma$) & 0.95\\
        value loss coefficient ($c_1$) & 0.1 \\
        entropy loss coefficient ($c_2$) & 0.02 \\
        learning rate & $10^{-4}$\\
        ppo-epoch & 4\\
        number-mini-batch & 7\\
        value-clip & 0.15\\
        policy-clip & 0.15\\
        buffer size ($\mathcal{B_I}$) & 10\\
        \bottomrule
    \end{tabular}
\vspace{1cm}
\end{minipage} \hspace{0.5cm} 
\begin{minipage}{0.45\textwidth}
    \centering
    \caption{PPO Hyper-parameters}
    \label{table2}
    \setlength{\tabcolsep}{16pt}
    \begin{tabular}{lcr}
        \toprule
        Parameter  & Values \\
        \midrule
        learning rate & $3e^{-4}$ \\
        n-steps   & 2048\\
        batch size & 64\\
        n-epochs     & 10 \\
        discount factor       & 0.99\\
        gae-gamma & 0.95\\
        clip-range & 0.2 \\
        normalize advantage & True \\
        vf-coef & 0.5\\
        max-grad-norm & 0.5\\
        \bottomrule
    \end{tabular}
\end{minipage}
\end{table}

\begin{figure*}[t]
\centering
\subfloat{\label{fig:im1}}
\subfloat{\label{fig:im2}}
\subfloat{\label{fig:im3}}
\subfloat{\label{fig:im4}}
\subfloat{\label{fig:im5}}
\begin{tikzpicture}
\node (im11) [xshift =-83ex,yshift = 8ex]{\includegraphics[scale=0.22]{leged.png}};
\node (im1) [xshift=-125ex, yshift = -5ex]{\includegraphics[scale=0.25]{PointMaze_Open_Diverse_GR-v3.pdf}};
\node (im2) [xshift=-98.5ex,yshift = -5ex]{\includegraphics[scale=0.25]{PointMaze_Medium_Diverse_G-v3.pdf}};
\node (im3) [xshift=-70.5ex,yshift = -5ex]{\includegraphics[scale=0.25]{PointMaze_Medium_Diverse_GR-v3.pdf}};
\node (im4) [xshift=-42.5ex, yshift = -5ex]{\includegraphics[scale=0.25]{PointMaze_Large_Diverse_G-v3.pdf}};
\node (im4) [xshift=-84ex, yshift = -28ex]{\includegraphics[scale=0.25]{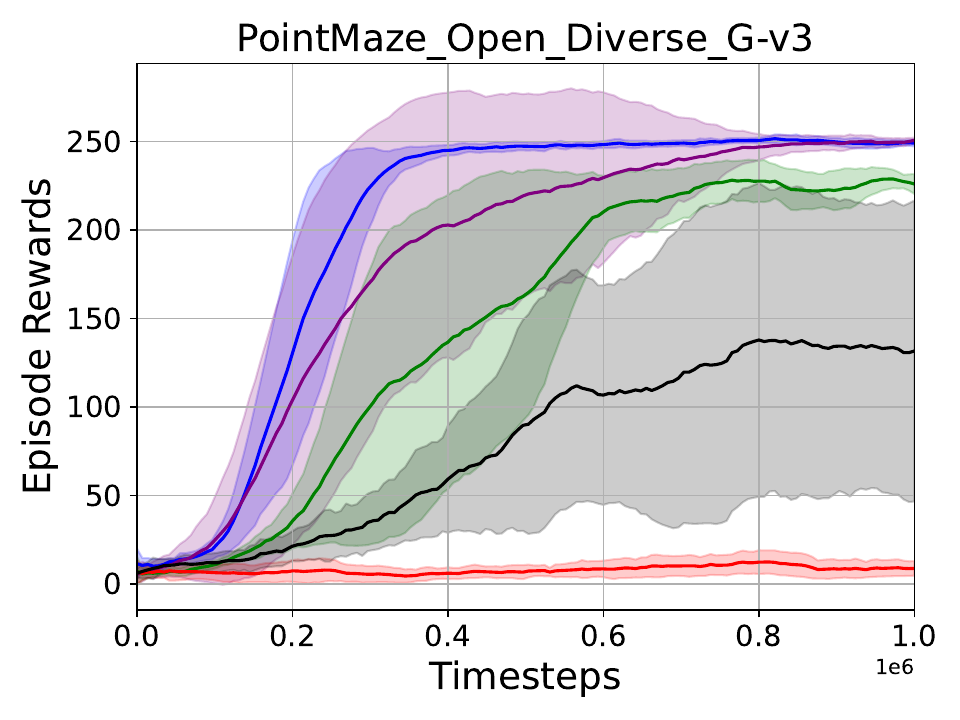}};

\end{tikzpicture}
\caption{Results show the performance on all 5 PointMaze multi-goal sparse reward tasks. The plots show the learning curves and the episodic rewards along the y-axis, evaluated under the current policy. The reported results are across seven different seeds. The proposed algorithms outperform all the baselines by a significant margin.}
\label{fig:maze_result}
\end{figure*}

\begin{figure*}[t]
\centering
\subfloat{\label{fig:im1}}
\subfloat{\label{fig:im2}}
\subfloat{\label{fig:im3}}
\subfloat{\label{fig:im4}}
\subfloat{\label{fig:im5}}
\begin{tikzpicture}
\node (im11) [xshift =-83ex,yshift = 8ex]{\includegraphics[scale=0.22]{legend2.png}};
\node (im1) [xshift=-125ex, yshift = -5ex]{\includegraphics[scale=0.25]{goal.pdf}};
\node (im2) [xshift=-98.5ex,yshift = -5ex]{\includegraphics[scale=0.25]{goal_behind_wall.pdf}};
\node (im3) [xshift=-70.5ex,yshift = -5ex]{\includegraphics[scale=0.25]{goal-tunnel.pdf}};
\node (im4) [xshift=-42.5ex, yshift = -5ex]{\includegraphics[scale=0.25]{reach_goal_occluded_tunnel_abe.pdf}};
\node (im5) [xshift=-84ex, yshift =-28ex]{\includegraphics[scale= 0.25]{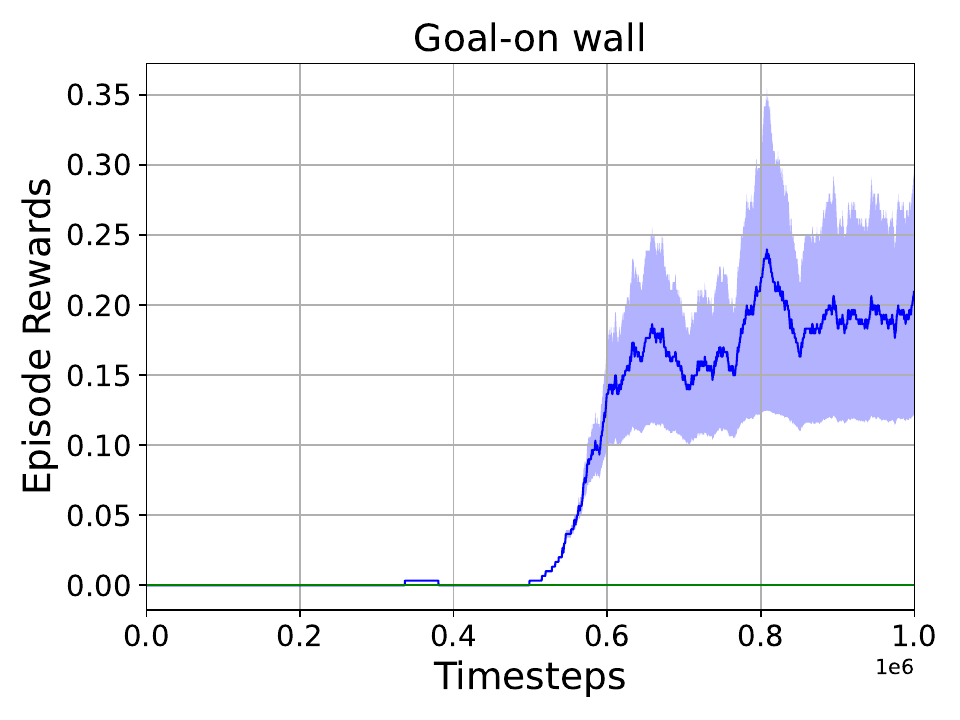}};

\end{tikzpicture}
\caption{Results show the performance on all 5 Animal-AI Olympics environments 
 sparse reward tasks. The plots show the learning curves, with episodic rewards (success rate) on the y-axis, evaluated under the current policy. The reported results are across 5 different seeds. The proposed algorithms outperform all the baselines by a significant margin.}
\label{fig:animal}
\end{figure*}

\begin{figure*}[t]
\centering
\subfloat{\label{fig:im11}}
\subfloat{\label{fig:im1}}
\subfloat{\label{fig:im2}}
\subfloat{\label{fig:im3}}
\subfloat{\label{fig:im4}}
\begin{tikzpicture}
\node (im11) [xshift =-83ex,yshift = 8ex]{\includegraphics[scale=0.22]{legend_abe.png}};
\node (im1) [xshift=-125ex, yshift = -5ex]{\includegraphics[scale=0.25]{MountainCarContinuous_abe.pdf}};
\node (im2) [xshift=-98.5ex,yshift = -5ex]{\includegraphics[scale=0.25]{Ant_abe.pdf}};
\node (im3) [xshift=-70.5ex,yshift = -5ex]{\includegraphics[scale=0.25]{HalfCheetah_abe.pdf}};
\node (im4) [xshift=-42.5ex, yshift = -5ex]{\includegraphics[scale=0.25]{Hopper_abe.pdf}};
\node (im6) [xshift=-125ex, yshift=-28ex]{\includegraphics[scale=0.25]{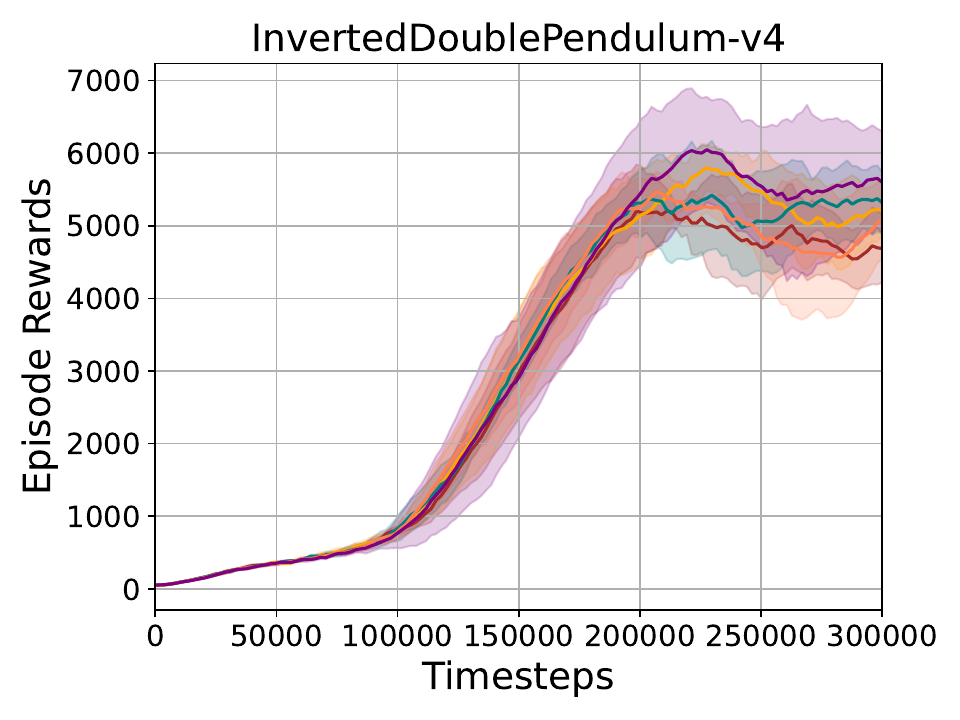}};
\node (im7) [xshift=-98.5ex, yshift=-28ex]{\includegraphics[scale=0.25]{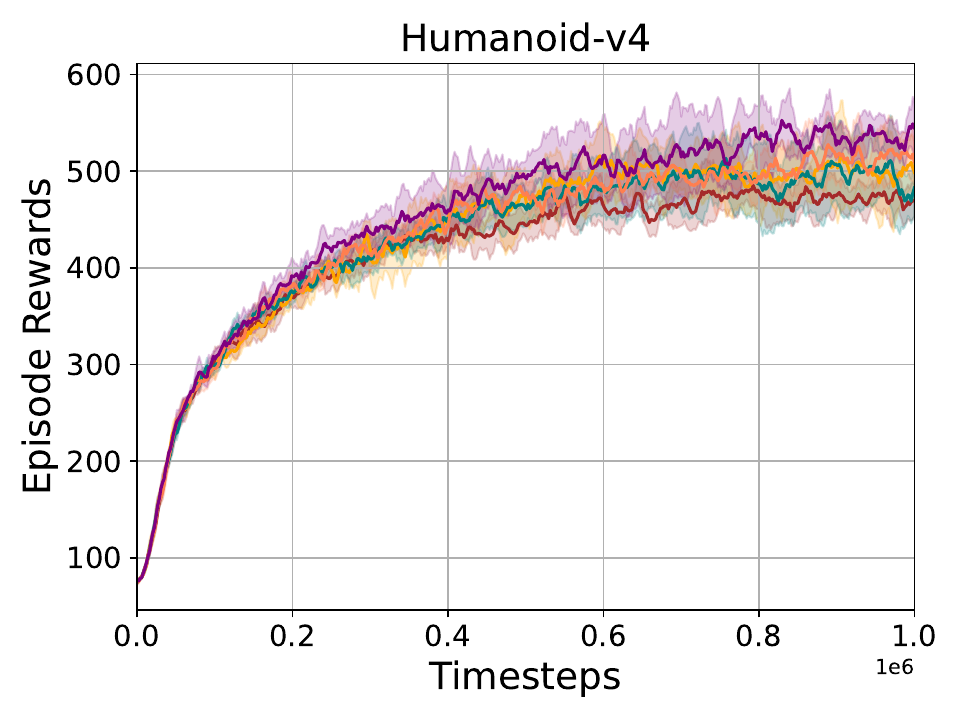}};
\node (im10) [xshift=-70.5ex, yshift=-28ex]{\includegraphics[scale=0.25]{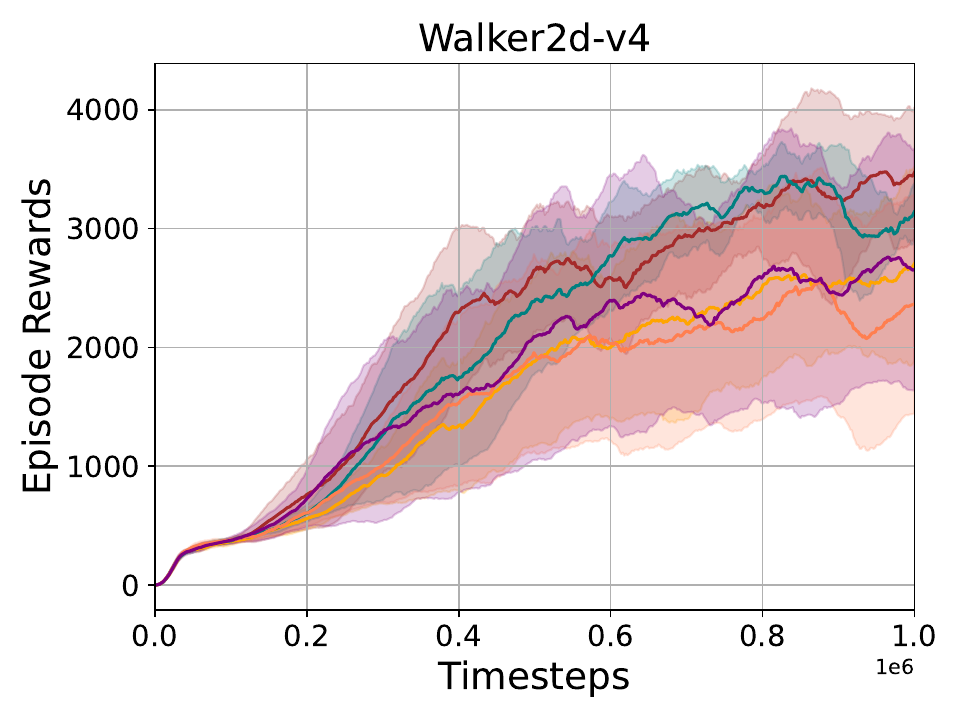}};
\node (im9) [xshift=-42.5ex, yshift=-28ex]{\includegraphics[scale=0.25]{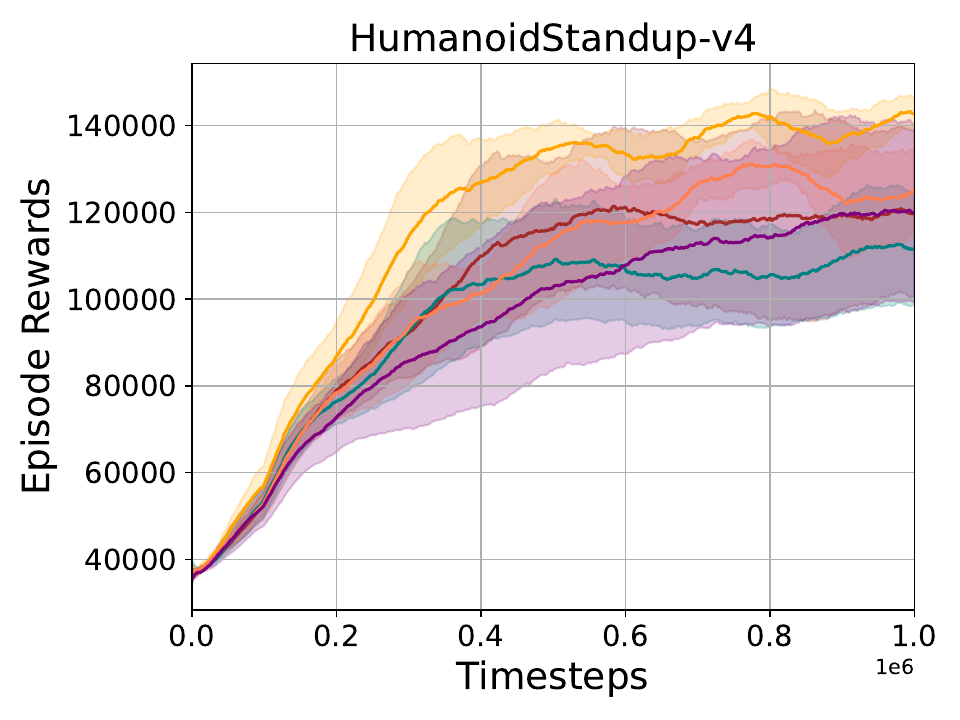}};


\end{tikzpicture}
\caption{Results show the ablation study on 8 MuJoCo~\citep{towers_gymnasium_2023} continuous control tasks. The parameter $\epsilon$ controls the balance between exploration and exploitation. The plots show the learning curves and the episodic rewards along the y-axis, evaluated under the current policy with different $\epsilon$. The reported results are the mean across five different seeds.}

\label{fig:match_abe}
\end{figure*}

\begin{figure*}[t]
\centering
\subfloat{\label{fig:im1}}
\subfloat{\label{fig:im2}}
\subfloat{\label{fig:im3}}
\subfloat{\label{fig:im4}}
\subfloat{\label{fig:im5}}
\begin{tikzpicture}
\node (im11) [xshift =-83ex,yshift = 8ex]{\includegraphics[scale=0.22]{legend_abe.png}};
\node (im1) [xshift=-125ex, yshift = -5ex]{\includegraphics[scale=0.25]{PointMaze_Open_Diverse_GR_abe.pdf}};
\node (im2) [xshift=-98.5ex,yshift = -5ex]{\includegraphics[scale=0.25]{PointMaze_Medium_Diverse_G_abe.pdf}};
\node (im3) [xshift=-70.5ex,yshift = -5ex]{\includegraphics[scale=0.25]{PointMaze_Medium_Diverse_GR_abe.pdf}};
\node (im4) [xshift=-42.5ex, yshift = -5ex]{\includegraphics[scale=0.25]{PointMaze_Large_Diverse_G_abe.pdf}};
\node (im4) [xshift=-84ex, yshift = -28ex]{\includegraphics[scale=0.25]{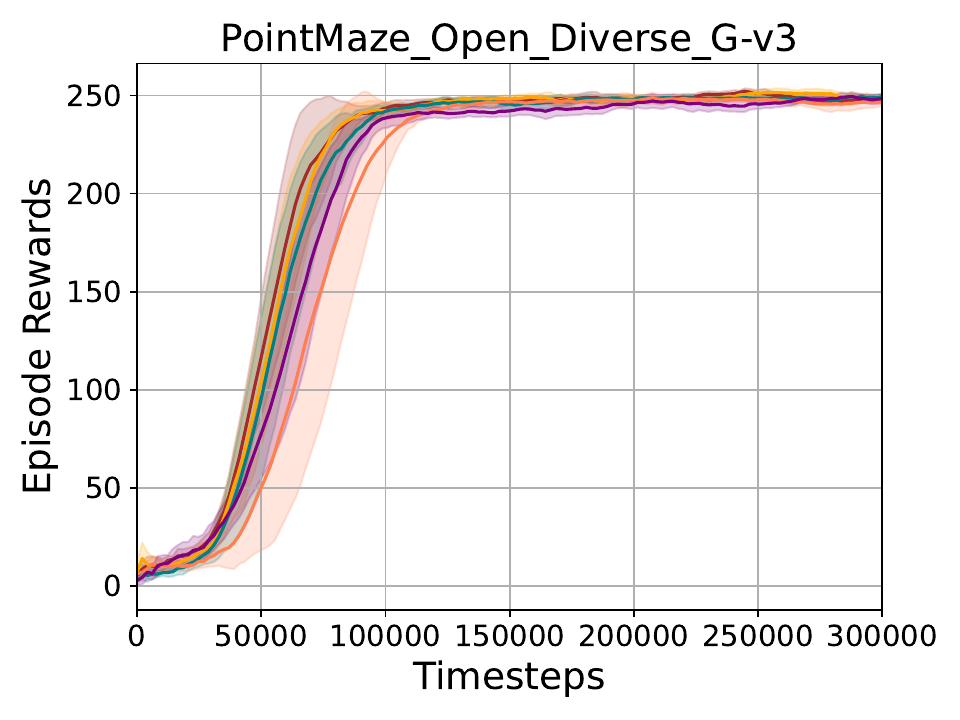}};
\end{tikzpicture}
\caption{Results show the ablation study on all 5 PointMaze multi-goal sparse reward tasks. The parameter $\epsilon$ controls the replay frequency to balance exploration vs exploitation. The plots show the learning curves and the episodic rewards along the y-axis, evaluated under the current policy with different $\epsilon$. The reported results are across five different seeds.}
\label{fig:maze_abe}
\end{figure*}

\end{document}